\documentclass[10pt,twocolumn,letterpaper]{article}
\usepackage[numbers,sort&compress]{natbib}

\usepackage[toc,page]{appendix}
\usepackage{iccv}
\usepackage{times}
\usepackage{epsfig}
\usepackage{graphicx}
\usepackage{amsmath}
\usepackage{amssymb}
\usepackage{float}
\usepackage{tabularx}
\usepackage{multirow}
\usepackage{makecell}
\usepackage{xspace}
\usepackage{siunitx}
\usepackage{booktabs}
\usepackage[accsupp]{axessibility}
\usepackage{mathtools, nccmath}

\usepackage[hypcap=false,font=small]{caption}
\usepackage{enumitem}

\usepackage[pagebackref=true,breaklinks=true,letterpaper=true,colorlinks,bookmarks=false]{hyperref}

\iccvfinalcopy %

\usepackage[capitalize]{cleveref}
\crefname{section}{Sec.}{Secs.}
\Crefname{section}{Section}{Sections}
\Crefname{table}{Table}{Tables}
\crefname{table}{Tab.}{Tabs.}

\newcommand{\PreserveBackslash}[1]{\let\temp=\\#1\let\\=\temp}
\newcolumntype{C}[1]{>{\PreserveBackslash\centering}p{#1}}
\newcolumntype{R}[1]{>{\PreserveBackslash\raggedleft}p{#1}}
\newcolumntype{L}[1]{>{\PreserveBackslash\raggedright}p{#1}}

\newcommand{\ours}{SHACIRA\xspace}

\newcommand{\Fig}{Fig.\xspace}
\newcommand{\myparagraph}[1]{\medskip\noindent\textbf{#1}}

\DeclareMathOperator\arctanh{\tanh^{-1}}
\def\Zb{\mathbf{Z}}
\def\fb{\mathbf{f}}
\def\Qb{\mathbf{Q}}
\def\Qhb{\widehat{\mathbf{Q}}}
\def\Qc{\lceil\Qhb\rceil}
\def\Qf{\lfloor\Qhb\rfloor}
\makeatletter
\DeclareRobustCommand\onedot{\futurelet\@let@token\@onedot}
\def\@onedot{\ifx\@let@token.\else.\null\fi\xspace}

\def\ie{\emph{i.e}\onedot}

\makeatother

\def\iccvPaperID{1898} %

\ificcvfinal\fi

\begin{document}

\title{%
SHACIRA: Scalable HAsh-grid Compression \\for Implicit Neural Representations
}

\author{Sharath Girish\\
University of Maryland\\
{\tt\small sgirish@cs.umd.edu}
\and
Abhinav Shrivastava\\
University of Maryland\\
{\tt\small abhinav@cs.umd.edu}
\and
Kamal Gupta\\
University of Maryland\\
{\tt\small kampta@cs.umd.edu}
}

\maketitle
\ificcvfinal\thispagestyle{empty}\fi

\begin{abstract}
\vspace{-3mm}
  Implicit Neural Representations (INR) or neural fields have emerged as a popular framework to encode multimedia signals such as images and radiance fields while retaining high-quality.
  Recently, learnable feature grids proposed by M\"uller et al.~\cite{muller2022instant} have allowed significant speed-up in the training as well as the sampling of INRs by replacing a large neural network with a multi-resolution look-up table of feature vectors and a much smaller neural network. However, these feature grids come at the expense of large memory consumption which can be a bottleneck for storage and streaming applications.
  In this work, we propose \ours, a simple yet effective task-agnostic framework for compressing such feature grids with no additional post-hoc pruning/quantization stages. We reparameterize feature grids with quantized latent weights and apply entropy regularization in the latent space to achieve high levels of compression across various domains. 
  Quantitative and qualitative results on diverse datasets consisting of images, videos, and radiance fields, show that our approach outperforms existing INR approaches without the need for any large datasets or domain-specific heuristics. Our project page is available at 
  \href{https://shacira.github.io}{https://shacira.github.io}.
\vspace{-5mm}
\end{abstract}

\begin{figure}[h]
\vspace{-1em}
\centering
\includegraphics[width=\linewidth]{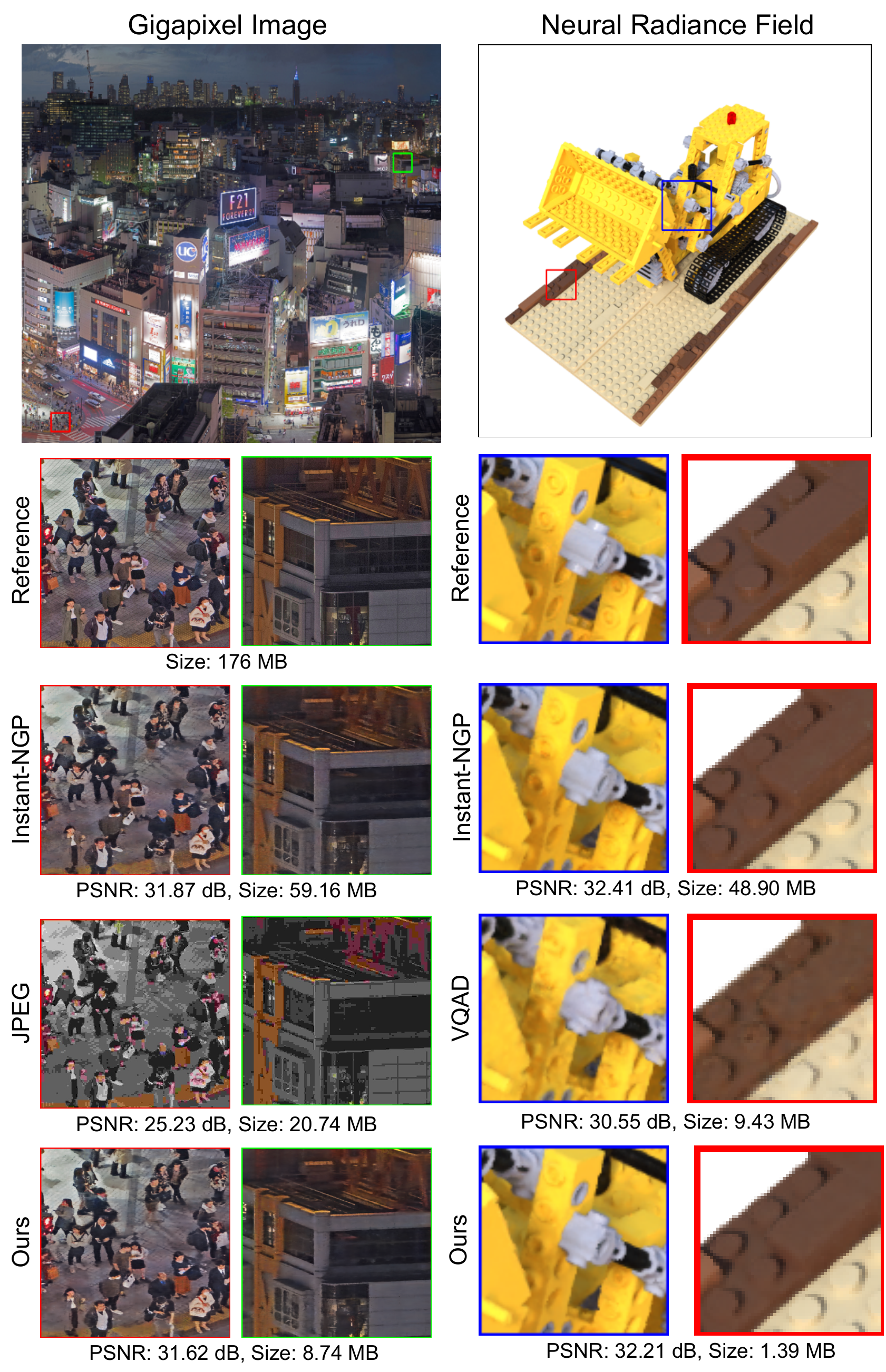}
\caption{We demonstrate the effectiveness of \ours for two tasks. The left column shows a gigapixel image at $21450 \times 56718$ resolution (cropped for visualization) encoded using Instant-NGP~\cite{muller2022instant}, JPEG2000~\cite{taubman2002jpeg2000}, and \ours (ours). The right column reconstructs NeRF~\cite{mildenhall2021nerf} from 2D images and their camera poses using Instant-NGP~\cite{muller2022instant}, VQAD~\cite{takikawa2022variable}, and \ours. For each example, we zoom into two crops to compare different methods. We show overall PSNR and size required by each method. \ours can capture high-resolution details with a smaller storage size in a task-agnostic way (only 2D/3D examples shown here).}
\label{fig:teaser}
\vspace{-3em}
\end{figure}

\section{Introduction}
In today's digital age, large quantities of data in different modalities (images, audio, video, 3D) is created and transmitted every day.
Compressing this data with minimal loss of information is hence an important problem and a number of techniques have been developed in the last few decades to address this challenging problem.
While the conventional methods such as JPEG~\cite{wallace1992jpeg} for images, HEVC~\cite{sze2014high} for videos excel at encoding signals in their respective domains, coordinate-based implicit neural representations (INR) or Neural Fields~\cite{10.1111:cgf.14505} have emerged as a popular alternative for representing complex signals because of their ability to capture high frequency details, and adaptability for diverse domains. INRs are typically multi layer perceptrons (MLPs) optimized to learn a scalar or vector field. They take a coordinate (location and/or time) as input and predict a continuous signal value(s) as output (such as pixel color/occupancy). Recently, various works have adapted INRs to represent a variety of signals such as audio~\cite{sitzmann2020implicit}, images~\cite{dupont2021coin,dupont2022coin++, tancik2021learned, mehta2021modulated}, videos~\cite{chen2021nerv,maiya2022nirvana}, shapes~\cite{park2019deepsdf,sitzmann2020metasdf}, and radiance fields~\cite{mildenhall2021nerf,liu2020neural}. Unsurprisingly, several methods have been proposed to compress INRs using quantization~\cite{strumpler2022implicit,dupont2022coin++,maiya2022nirvana}, pruning, or a combination of both~\cite{chen2021nerv}. The focus of these works is to compress the weights of the MLP, which often leads to either a big drop in the reconstruction quality, or slow convergence for high resolution signals.

In this work, we consider a different class of INR approaches that employ learnable multi-resolution feature grids~\cite{takikawa2021neural,muller2022instant}. These feature grids store feature vectors at different coordinate locations with varying Level-Of-Detail (LOD). The features from different levels (or resolutions) are concatenated and passed through a tiny MLP to reconstruct the required signal. This shifts the burden of representing the signal to the feature grid instead of the MLP. Such methods have shown to be effective in approximating complex signals such as 3D scenes and gigapixel images with high fidelity~\cite{muller2022instant} and fast training time (since the cost of lookup is very small). However, the size of the feature grids can be very large which is not memory efficient and impractical for many real-world applications with network bandwidth or storage constraints.

We propose an end-to-end learning framework for compressing such feature grids without any loss in reconstruction performance. Our feature grid consists of quantized feature vectors and parameterized decoders which transform the feature vectors into continuous values before passing them to MLP. We use an entropy regularization loss on the latent features to reduce the size of the discrete latents without significantly affecting the reconstruction performance. To address the discretization gap inherent to this discrete optimization problem, we employ an annealing approach to the discrete latents which improves the training stability of the latents, converging to better minima. Both entropy regularization and reconsutruction objective can be trained jointly in an end-to-end manner without requiring post-hoc quantization, pruning or finetuning stages. Further, the hierarchical nature of feature grids allows scaling to high dimensional signals unlike pure MLP-based implicit methods.

As seen in \Cref{fig:teaser}, the proposed approach is able to compress feature-grid methods such as Instant-NGP~\cite{muller2022instant} with almost an order of magnitude while retaining the performance in terms of PSNR for gigapixel images and 3D scenes from the RTMV dataset~\cite{tremblay2022rtmv}. We also conduct extensive quantitative experiments and show results on standard image compression benchmarks such as the Kodak dataset outperforming the classic JPEG codec as well as other implicit methods in the high compression regime. Our approach can even be trivially extend to videos, performing competitively with video-specific INR methods such as NeRV~\cite{chen2021nerv}, without explicitly exploiting the inherent temporal redundancy present in videos. The key contribution of our work is to directly compress the learnable feature grid with proposed entropy regularization loss and highlight its adaptibility to diverse signals. We summarize our contributions below:

\begin{itemize}[leftmargin=*,itemsep=0em]
    \item We introduce an end-to-end trainable compression framework for implicit feature grids by maintaining discrete latent representations and parameterized decoders.
    \item We provide extensive experiments on compression benchmarks for a variety of domains such as images, videos, and 3D scenes showing the generalizability of our approach 
    outperforming a variety of INR works.
\end{itemize}

\begin{figure*}[h]
\centering
\includegraphics[height=0.44\textwidth]{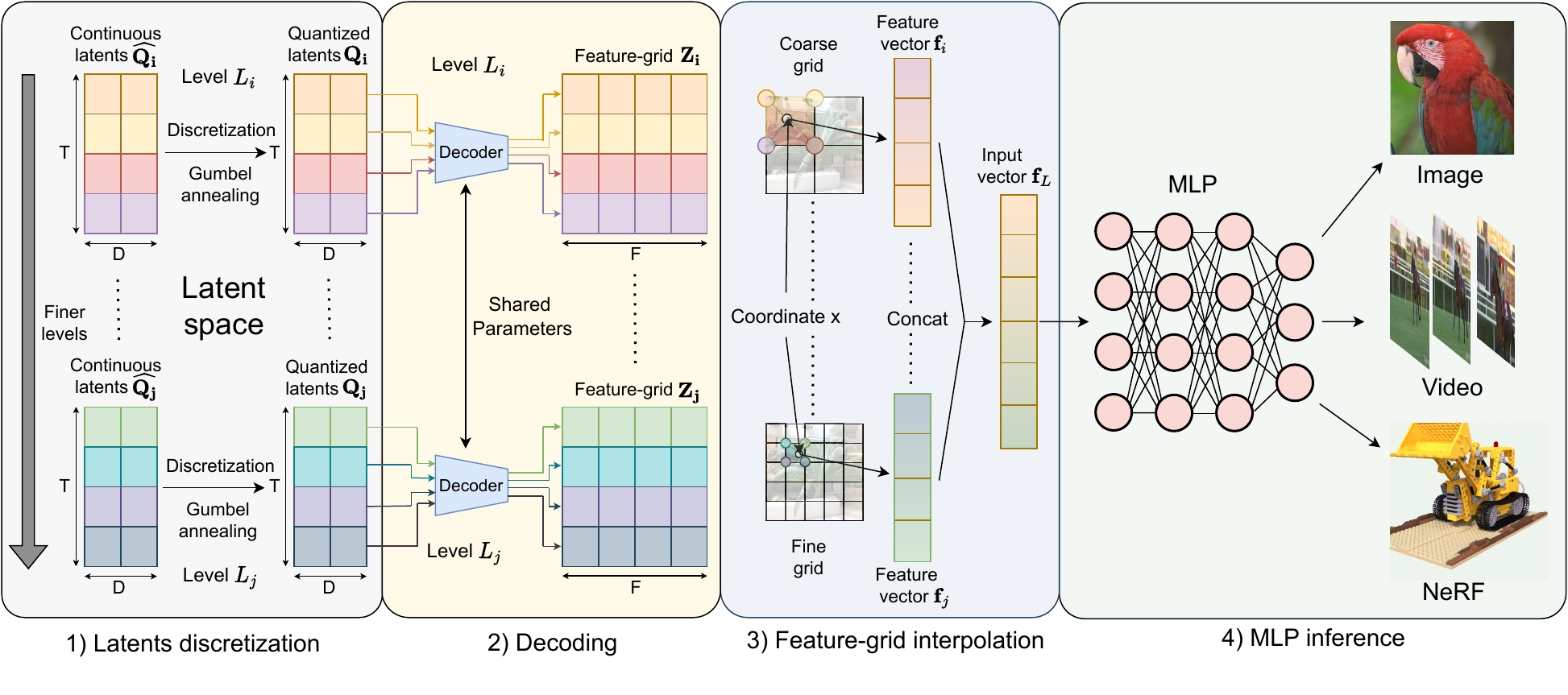}
\caption{Overview of our approach: We maintain latent representations which are quantized and decoded using parameterized decoders to obtain a hash table/feature-grid at different levels. We then index the input coordinate into the hash table to obtain feature vectors. The feature vectors are then concatenated and passed through an MLP to obtain the output signal.}
\label{fig:approach}
\end{figure*}
\section{Related work}

\noindent\textbf{Learned image/video compression:} A large number of neural compression works for images consist of an autoencoder framework~\cite{balle2016end,theis2017lossy} which transform/encode a data point to a latent code and decode the quantized latent to obtain a reconstruction of the data signal. The autoencoder is typically trained in an end-to-end fashion on a large training dataset by minimizing a rate-distortion objective. Numerous extensions to these works introduce other improvements such as hyperpriors~\cite{balle2018variational}, autoregressive modeling~\cite{minnen2018joint}, Gaussian mixture likelihoods and attention modules~\cite{cheng2020learned}, improved inference~\cite{yang2020improving}. Another set of works extends this framework for videos as well~\cite{lu2019dvc,habibian2019video,agustsson2020scale,yang2020learning} exploiting the inherent temporal redundancy. These approaches achieve impressive compression results outperforming classical codecs in their respective domains. In contrast, we focus on a different class of works involving implicit neural representations (INRs) which overfit a network to each datapoint and store only the network weights to achieve data compression.

\myparagraph{INRs and application to data compression:} INRs~\cite{stanley2007compositional} are a rapidly growing field popularized for representing 3D geometry and appearance~\cite{park2019deepsdf,mildenhall2021nerf,chen2019learning,mescheder2019occupancy} and have since been applied to a wide variety of fields such as GANs~\cite{chen2021learning,karras2021alias}, image/video compression~\cite{dupont2021coin,dupont2022coin++,chen2021nerv,maiya2022nirvana,strumpler2022implicit}, robotics~\cite{simeonov2022neural} and so on. Meta-learning on auxiliary datasets has been shown to provide good initializations and improvements in reconstruction performance while also greatly increasing convergence speed for INRs~\cite{sitzmann2020metasdf,tancik2021learned,strumpler2022implicit}. 
Our approach can similarly benefit from such meta-learning stages but we do not focus on it and rather highlight the effectiveness of our approach to compress feature-grid based INRs and its advantages over MLP-based INRs. Perhaps the closest work to our approach is that of VQAD~\cite{takikawa2022variable} which performs a vector quantization of these feature grids learning a codebook/dictionary and its mapping to the feature grid in 3D domain. They however learn a fixed-size codebook without any regularization loss and do not perform well for high-fidelity reconstructions as we discuss in \Cref{sec:experiments}.

\myparagraph{Model compression:} As INRs represent data as neural networks, they transform the data compression problem to a model compression one. Many works exist for model compression involving pruning for achieving high levels of sparsity~\cite{frankle2018lottery,frankle2019stabilizing,girish2021lottery,gale2019state} or quantization for reducing the number of bits necessary~\cite{lin2017towards,tung2018clip,fan2020training}. Another line of works~\cite{oktay2019scalable,Girish2022LilNetXLN} perform compression similar to~\cite{balle2018variational} using quantized latents with entropy regularization losses. These methods, however, are specific to convolutional networks and are not trivially extensible to compress INRs.

\section{Approach}
Our goal is to simultaneously train and compress feature-grid based implicit neural networks. \Cref{ssec:feature_grid} provides a brief overview of feature-grid INRs proposed in \cite{muller2022instant}. \Cref{ssec:grid_reparam} describes various components of our approach while \Cref{ssec:grid_compression} discusses compressing feature grids. Our approach for end-to-end feature grid compression is outlined in \Cref{ssec:end_to_end} and also illustrated in \Cref{fig:approach}.

\subsection{Feature-grid INRs}
\label{ssec:feature_grid}
INRs or neural fields~\cite{10.1111:cgf.14505} typically learn a mapping $g_{\phi}(\mathbf{x}): \mathbb{R}^d\rightarrow~\mathbb{R}^c$ where $g$ is an MLP with parameters $\phi$. Input $\mathbf{x} \in \mathbb{R}^d$ to the MLP are $d$ dimensional coordinates, where $d=2$ in the case of images, $3$ in the case of videos, or $5$ in the case of radiance fields. $c$-dimensional output can represent RGB colors or occupancy in space-time for the given input coordinates. Such methods are able to achieve high-quality reconstructions, however, suffer from long training times and scalability to high-resolution signals. A few works have suggested utilizing fixed positional encodings of input coordinates~\cite{tancik2020fourier} to be able to reconstruct complex high-resolution signals but the training time of these networks still poses a big challenge. To alleviate this issue, \cite{muller2022instant,takikawa2021neural} proposed storing the input encodings in the form of a learnable feature grid. The feature grid allows for significant speed-up in the training time by replacing a large neural network with a multi-resolution look-up table of feature vectors and a much smaller neural network. We now provide a brief overview of Instant-NGP~\cite{muller2022instant}.

For ease of explanation, for the rest of this section, we will assume input is a 2D signal $\mathbf{x}\in\mathbb{R}^2$. However, all the techniques we discuss can be directly applied to the 3D case. In this framework, we represent the feature grid by a set of parametric embeddings $\Zb$. $\Zb$ is arranged into $L$ levels representing varying resolutions or Levels-Of-Detail (LOD). More formerly, each level has its own embedding matrix $\Zb_l$, and hence $\Zb = \{\Zb_1, \dots,\Zb_l\}$. The number of feature vectors in the embedding matrix $\Zb_l$ depends on the resolution of the level $R_l$. For coarser resolutions, we allow $\Zb_l$ to consist of $R_l^2$ rows, but for finer resolutions, we cap the maximum number of rows in the matrix to $T$. Hence for $\mathbf{x}\in\mathbb{R}^2$, 
\begin{equation}
    \Zb_l\in\mathbb{R}^{T_l\times F}, \textnormal{ where } T_l = \textnormal{min}(R_l^2,T)
\end{equation}
Here $F$ is the dimension of feature vectors and is kept fixed for all levels.
For a given input $\mathbf{x}$, we can obtain the 4 closest corner indices $\{tl,tr,bl,br\}$ within the $R_l\times R_l$ grid. Each of the corner index maps to an entry in $\Zb_l$. This mapping is direct when $R_l^2\leq T$ and uses a hashing function~\cite{teschner2003optimized} otherwise. The feature vector $\fb_l$ at level $l$ for the input $\mathbf{x}$ is then obtained by a simple bilinear interpolation of feature vectors of the corner indices, \ie 
\begin{equation}
    \fb_l(\mathbf{x})=\text{interp}(\Zb_l[tl],\Zb_l[tr],\Zb_l[bl],\Zb_l[br])
\end{equation}
Note that in the case of 3D input, we consider 8 closest corner indices, and perform a trilinear interpolation. $\fb_l(\mathbf{x})$ is concatenated across different levels to obtain the overall feature vector $\fb(\mathbf{x})\in\mathbb{R}^{FL}$, which is passed as input to the neural network $g_\phi$.
\begin{equation}
    \widehat{\textbf{y}} = g_\phi(\text{concat}[\fb_1(\mathbf{x}),\dots,\fb_L(\mathbf{x})])
\end{equation}
Here $\widehat{\textbf{y}}$ is the final prediction of INR for the input $\mathbf{x}$.
Since the concatenation, indexing, and interpolation operations are differentiable, parameters $\{\phi,\Zb\}$ can be optimized using any gradient based optimizer by minimizing a loss function $\mathcal{L}(\widehat{\textbf{y}},\textbf{y})$ between the predicted $\widehat{\textbf{y}}$ and ground-truth signal $\textbf{y}$. $\mathcal{L}$ can be any differentiable loss function such as the Mean Squared Error (MSE). The MLP is typically very small in comparison to MLP-based INRs. Thus, $\phi$ consists of far fewer parameters than $\Zb$. Such an INR design allows for much faster training and inference as the cost of indexing into the feature grid is quite small.

\subsection{Feature-grid reparameterization}
\label{ssec:grid_reparam}
While feature-grid INRs can converge faster than pure-MLP approaches, the space required to store all the parameters at different levels can rise very rapidly if we want high-fidelity reconstructions for high-resolution inputs. This makes them unsuitable for resource-constrained applications. We thus aim to reduce the storage size of these feature grids. To this effect, we propose to maintain discrete or quantized latent representations $\Qb_l\in\mathbb{R}^{T_l\times D}$ for each embedding $\Zb_l \in\mathbb{R}^{T_l\times F}$. The latents, consisting of only integer values, can be of any dimension $D$ with a larger $D$ allowing for greater representation power at the cost of storage size. 

In order to map these discrete latent features $\Qb_l$ to the continuous features in the embedding table $\Zb_l$, we propose a parameterized decoder $d_\theta:\mathbb{R}^D\rightarrow\mathbb{R}^F$. While it is possible to use separate complex decoders for each level,
in practice, we observed that a single shared decoder parameterized as a linear transform across all L levels works pretty well 
and has a minimal impact on the training/inference times and fidelity of reconstructions.

Note that the quantized latents $\Qb$ are no more differentiable (here we dropped the subscript $l$ without loss of generality for notational simplicity). In order to optimize these quantized latents, we maintain a continuous proxy parameters $\Qhb$ of the same size as $\Qb$. $\Qb$ is obtained by rounding $\Qhb$ to the nearest integer. To make this operation differentiable, we utilize the Straight-Through Estimator~\cite{bengio2013estimating}(STE). In STE, we use quantized weights $\Qb$ during the forward pass, but use the continuous proxies to propagate the gradients from $\Qb$ to $\Qhb$ during the backward pass.

STE serves as a simple differentiable approximation to the rounding operation but leads to a large rounding error $\|\Qb-\Qhb\|$. To overcome this issue, we utilize an annealing approach~\cite{yang2020improving} to perform a soft rounding operation. We represent $\Qb$ by either rounding up (denoted as $\lceil.\rceil$) or down (denoted as $\lfloor.\rfloor$) $\Qhb$ to the nearest integer. Using one-hot gates $b \in \{0,1\}$, where $b=0$ corresponds to rounding up, $b=1$ corresponds to rounding down, 
we can represent $\Qb = b\Qf + (1-b)\Qc$.
The gate $b$ is sampled from a soft relaxed 2-class distribution
\begin{align}
    \text{Prob}(b=0) \propto \exp\left\{ -\arctanh{\left(\Qhb-\Qf\right)}/\tau \right\} \nonumber \\
    \text{Prob}(b=1) \propto \exp\left\{ -\arctanh{\left(\Qc-\Qhb\right)}/\tau \right\}
\end{align}

where $\tau>0$ represents the temperature parameter. $\Qhb$ approaching either $\lfloor\Qhb\rfloor$ or $\lceil\Qhb\rceil$ thus increases the likelihood of sampling to that respective value. In the beginning of the training, $\tau=1$ and is a poor approximator of rounding operation but provides more stable gradients. As training progresses, $\tau$ is annealed towards zero, so that the random distribution converges to a deterministic one at the end of training. The gradients are propagated through the samples using the Gumbel reparameterization trick~\cite{jang2016categorical}.

\subsection{Feature-grid compression}
\label{ssec:grid_compression}

To further improve compression levels, we minimize the entropy of the latents using learnable probability models~\cite{balle2018variational}. We note that our learned latents $\Qb$ or $\Qhb$ are of dimensions $T\times D$. We can interpret these latents as consisting of $T$ samples from a discrete $D$ dimensional probability distribution. 
We introduce $D$ probability models one for each latent dimension $d\in\{1,\dots,D\}$, $P_d:\mathbb{R}\rightarrow[0,1]$. We discuss the exact form of $P_d$ in the supplementary material.
With these probability models, we can minimize the length of a bit sequence encoding $\Qhb$ by minimizing the self-information loss or the entropy of $\Qhb$~\cite{shannon1948mathematical}
\begin{equation}
    \mathcal{L}_I(\Qhb) = -\frac{1}{T}\sum_{d=1}^{D}\sum_{i=1}^{T}\textnormal{log}_2\left(P_d\left(\Qhb[i,d]+n\right)\right)
    \label{eq:inf_loss}
\end{equation}
where $n\sim\mathcal{U}[-1,1]$ represents the uniform random distribution to approximate the effects of quantization, and $\mathcal{L}_I$ is the self-information loss.

\subsection{End-to-end optimization}
We provide an overview of our approach in \cref{fig:approach}. We maintain continuous learnable latents $\Qhb$. The proposed annealing approach (\cref{ssec:feature_grid}) progressively converges the continuous $\Qhb$ to the discrete $\Qb$. The approach is made differentiable using the gumbel reparameterization trick and straight-through estimator. $\Qb$ is passed through the decoder $d_\theta$ with parameters $\theta$ to obtain the feature grid table $\Zb$. We then index $\Zb$ at different levels/resolutions using the coordinates $\textbf{x}$ to obtain a concatenated feature vector $\fb$ which is passed through an MLP $g_\phi$ to obtain the predicted signal $\widehat{\textbf{y}}$. 
\begin{align}
    \Qb &= \text{discretize}(\Qhb) \\
    \Zb &= d_\theta(\Qb) \\
    \widehat{\mathbf{y}} &= g_\phi\left( \text{concat}\left[\text{interp}(\Zb, \mathbf{x})\right]\right)
\end{align}
Our framework is thus fully differentiable in the forward and backward pass. For a given signal $\mathbf{y}$ and its corresponding input coordinate grid $\mathbf{x}$, we optimize the parameters $\phi$ of MLP $g_\phi$, discrete feature grid $\Qhb$, discrete to continuous decoder $\theta$, and the probability models $\{P_d\}$ in an end-to-end manner by minimizing the following rate distortion objective 
\begin{equation}
    \mathcal{L}_\textnormal{MSE}(\widehat{\mathbf{y}},\textbf{y})+\lambda_I\mathcal{L}_I(\Qhb)
    \label{eq:net_loss}
\end{equation}
where $\lambda_I$ controls the rate-distortion optimization trade-off. Post training, the quantized latents are stored losslessly using algorithms such as arithmetic coding~\cite{witten1987arithmetic} utilizing the probability tables from the density models $\{P_d\}$.
\label{ssec:end_to_end}
\section{Experiments}
\label{sec:experiments}
We apply our INR framework to images, videos, and radiance fields. We outline our experimental setup in \cref{ssec:exp_setup}.\ \cref{ssec:img_compress}, \cref{ssec:3d_compress}, and \cref{ssec:video_compress} provide results on compression for images, radiance fields, and videos respectively.\ 
 \cref{ssec:streaming} illustrates the application of our approach for progressive streaming. \cref{ssec:convergence} discusses the convergence speeds of our approach. \cref{ssec:ent_anneal} ablates the effect of entropy regularization and annealing. Additional experiments and ablations are provided in the supplementary material.

\subsection{Experimental details and setup}
\label{ssec:exp_setup}
We show image compression performance on the Kodak dataset~\cite{kodak} consisting of 24 images of resolution $512\times768$. To show our scalability to larger images, we provide results on higher resolution images varying from 2M pixels to 1G pixels. We use the Kaolin-Wisp library~\cite{KaolinWispLibrary} for our experiments on radiance fields. In addition to the Lego scene in \cref{fig:teaser}, we provide results on 10 bricks scenes from the RTMV dataset~\cite{tremblay2022rtmv} which contains a wide variety of complex 3D scenes. For videos, we benchmark on the UVG dataset~\cite{mercat2020uvg} consisting of seven 1080p resolution videos consisting of 600 frames each at 120fps. Additionally, we extract the first frame from these 7 videos to create a 7-image dataset, UVG-F, for benchmarking image compression at $1080\times1920$ resolution against other popular INR compression approaches. We primarily measure distortion in terms of the PNSR (dB) and rate in terms of Bits-Per-Pixel (BPP) (or model size for neural fields).

We fix the minimum grid resolution to 16 and vary the number of per-level entries T, the number of levels L, and the maximum grid resolution for various modalities to control the rate/distortion trade-off. We fix the number of hidden MLP layers to 2 with the ReLU activation. We fix the batch size (number of coordinates per iteration) to be $2^{18}$ for all our experiments except Kodak where we pass the full coordinate grid. The entropy regularization parameter is set to $1.0e^{-4}$ for all our experiments unless specified otherwise. The parameters are optimized using the Adam optimizer~\cite{kingma2014adam}. We provide additional experimental details in the supplementary material.

\begin{figure}
    \centering
    \includegraphics[width=1.0\linewidth]{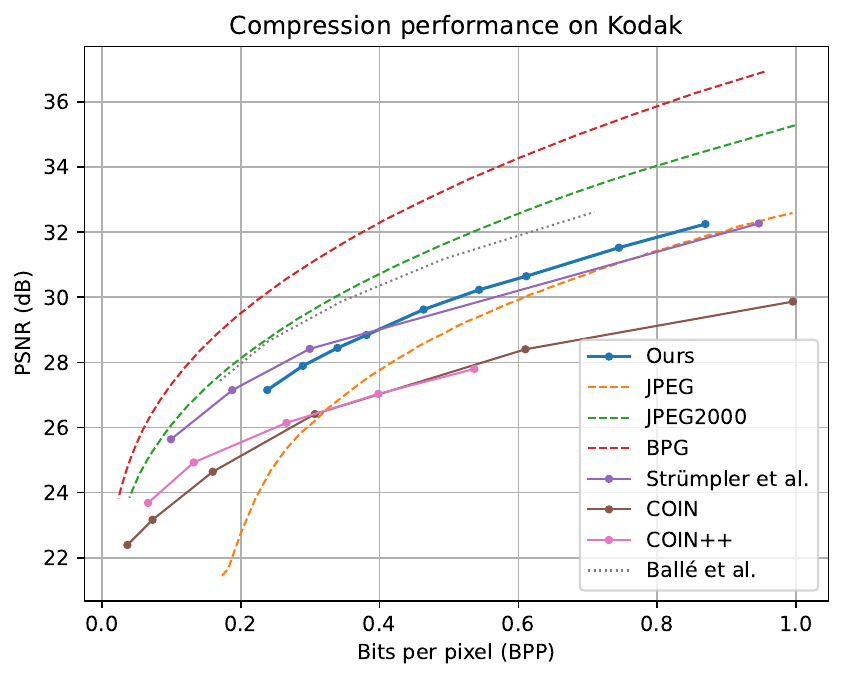}
    \caption{Comparison of our approach on the Kodak image dataset with classical (dashed), RDAE (dotted), INR (solid) methods. We outperform state-of-the-art INR approaches bridging the gap to classical and RDAE methods.}
    \label{fig:kodak_baselines}
\end{figure}
\renewcommand{\arraystretch}{1.1}
\begin{table}[t]
\caption{Image compression at varying resolutions. We compare against implicit network methods of INGP~\cite{muller2022instant}, Positional~\cite{strumpler2022implicit}, SIREN~\cite{sitzmann2020implicit}, and the auto-encoder based work, RDAE~\cite{balle2018variational}.
We achieve high values of PSNR, maintaining similar quality reconstructions as INGP while requiring far fewer bits ($4-9{\times}$).}
\centering
\resizebox{\linewidth}{!}{
\begin{tabular}{@{}L{\dimexpr.25\linewidth}C{\dimexpr.22\linewidth}C{\dimexpr.15\linewidth}C{\dimexpr.14\linewidth}C{\dimexpr.11\linewidth}@{}}
\toprule
Image & Method & PSNR $\uparrow$ & SSIM $\uparrow$ & BPP $\downarrow$ \\
\midrule
\multirow{4}{*}{\makecell{UVG-F \\($1920\times 1080$)}} & Positional~\cite{strumpler2022implicit} & 33.17&0.86&1.52\\
 & JPEG~\cite{wallace1992jpeg} & 36.98& 0.91 &0.76\\
 & RDAE~\cite{balle2018variational}& 34.23 & \textbf{0.93} & 0.76\\
  & Ours & \textbf{37.74} &0.92&\textbf{0.76}\\
\midrule
\multirow{4}{*}{\makecell{SMACS \\($4630\times 4537$)}} & INGP~\cite{muller2022instant} & 34.61 &0.86&0.18\\
 & JPEG~\cite{wallace1992jpeg} & 34.77 &0.86&0.18\\
 & RDAE~\cite{balle2018variational} & 34.06 &\textbf{0.89}&0.40\\
  & Ours & \textbf{34.90} &0.86&\textbf{0.04}\\
\midrule
\multirow{5}{*}{\makecell{Cosmic-Cliffs \\($8441\times 14575$)}} & INGP~\cite{muller2022instant} & 38.78 &0.96&0.63\\
 & SIREN~\cite{sitzmann2020implicit} & 27.32 &0.90&0.14\\
 & JPEG~\cite{wallace1992jpeg} & 38.38 &0.95&0.29\\
 & RDAE~\cite{balle2018variational} & 37.90&\textbf{0.97}&0.38\\
  & Ours & \textbf{38.79} &0.96&\textbf{0.11}\\
\midrule
\multirow{3}{*}{\makecell{Pearl \\($23466\times 20000$)}} & INGP~\cite{muller2022instant} & \textbf{29.62} &0.84&1.00\\
 & JPEG~\cite{wallace1992jpeg} & 29.10 &0.84&0.29\\
  & Ours & 29.44 &\textbf{0.84}&\textbf{0.12}\\
\midrule
\multirow{3}{*}{\makecell{Tokyo \\($21450\times 56718$)}} & INGP~\cite{muller2022instant} & \textbf{31.87} &\textbf{0.82}&0.39\\
 & JPEG~\cite{wallace1992jpeg} & 31.16 &0.82&0.18\\
  & Ours & 31.62 &0.82&\textbf{0.06}\\

\bottomrule
\end{tabular}
} %
\vspace{-1em}
\label{tab:gigapix_baselines}
\end{table}

\begin{figure}[t]
\centering
   \includegraphics[width=\linewidth]{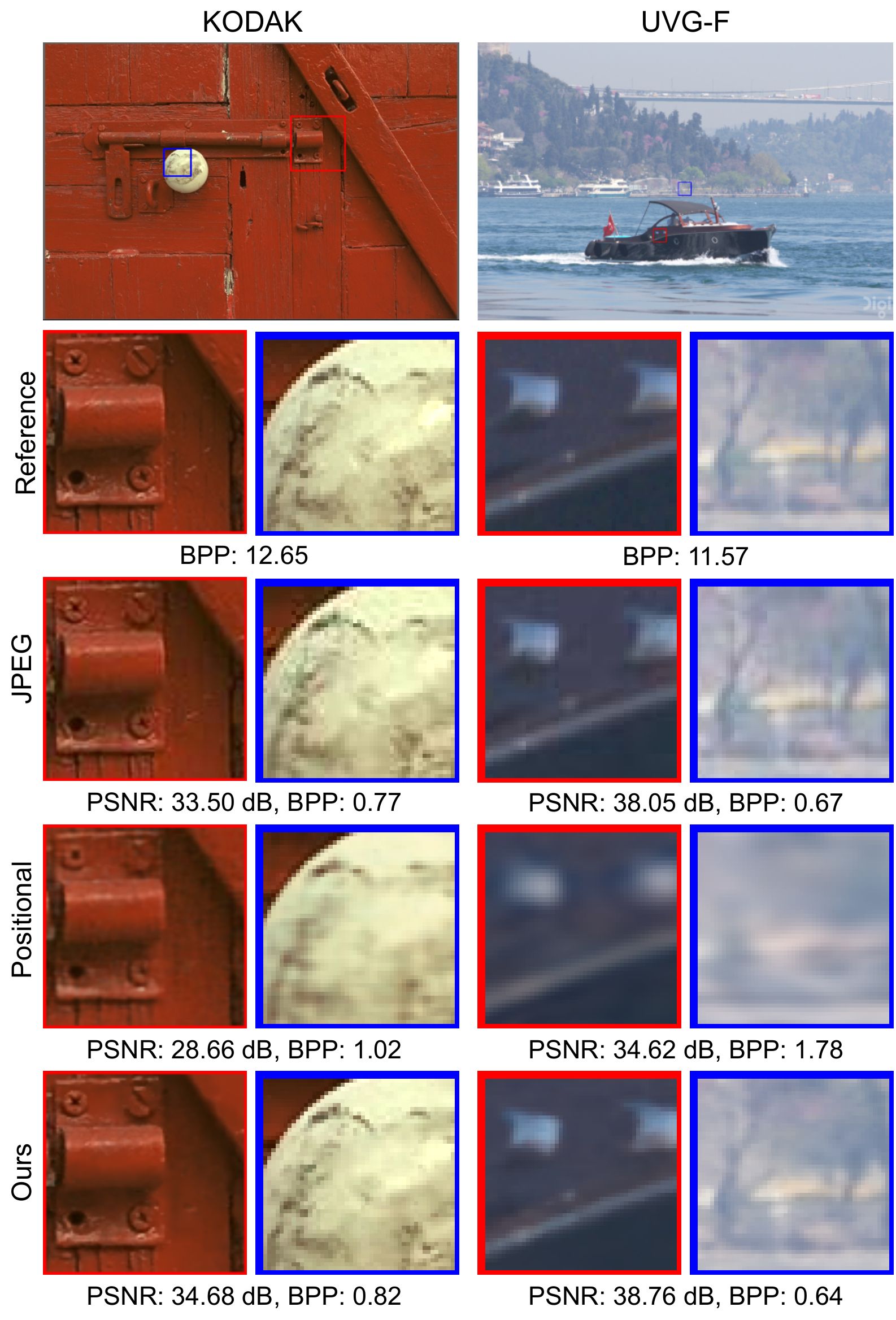}
   \caption{Qualitative results on Kodak and UVG-F: We obtain much higher quality reconstructions capturing finer detail compared to~\cite{strumpler2022implicit} or JPEG. Notice the blocking and discoloration artifacts present in JPEG which are significant at lower BPP values.}
   \vspace{-1em}
\label{fig:kodak_qualitative}
\end{figure}
\begin{figure*}[t]
\centering
   \includegraphics[width=0.95\linewidth]{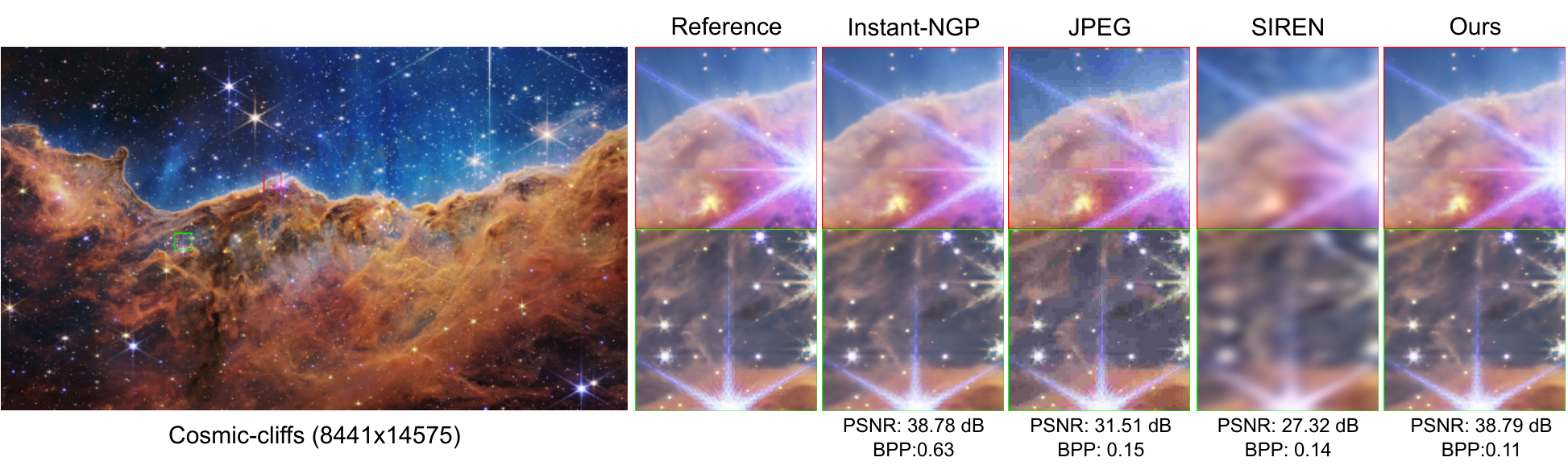}
   \vspace{-1em}
   \caption{Compression result visualization on the Cosmic-Cliffs image for 4 methods. We obtain similar PSNR and reconstruction quality as Instant-NGP while ${\sim}6{\times}$ smaller. SIREN fails to fit high frequency information leading to blurry patches as seen. JPEG on the other hand suffers from blocking artifacts and discoloration leading to drop in reconstruction quality.}
\label{fig:cosmic}
\end{figure*}

\begin{figure*}[t]
\centering
   \includegraphics[width=0.95\linewidth]{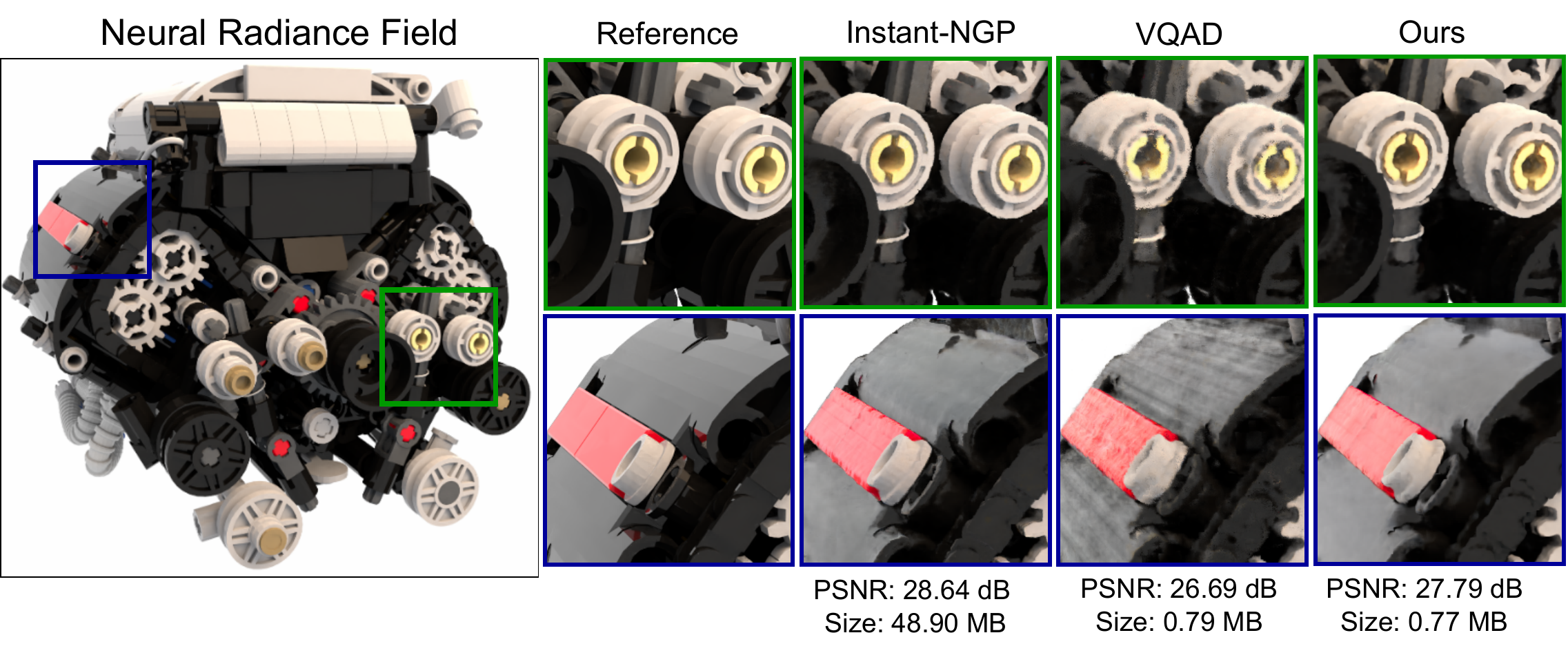}
   \vspace{-1em}
   \caption{Evaluation on V8 from the RTMV dataset ($1600\times1600$ resolution). We obtain ${\sim}60{\times}$ compression compared to Instant-NGP but with a PSNR drop of $1dB$. We outperform VQAD obtaining higher PSNR at similar model size. We also obtain much finer reconstructions when compared with VQAD as shown in the zoomed patches.}
\label{fig:v8_qualitative}
\end{figure*}

\subsection{Scalable image compression}
\label{ssec:img_compress}
We visualize results on the Kodak benchmark in \cref{fig:kodak_baselines}. We outperform the MLP-based INR approaches of COIN~\cite{dupont2021coin}, COIN++~\cite{dupont2022coin++} by a significant margin at all bitrates. We outperform~\cite{strumpler2022implicit}, which utilizes positional encodings and weight quantization, at higher bitrates while also having much lesser encoding times (as we show in \cref{ssec:convergence}) %
We show qualitative results on one of the images from the KODAK dataset in \cref{fig:kodak_qualitative}. We obtain higher quality reconstructions ($28.66\rightarrow34.68$ PSNR) capturing fine details while also requiring fewer bits for storage ($1.02\rightarrow0.88$ BPP) as compared to~\cite{strumpler2022implicit}. We however, observe slightly lower performance in the low BPP regime where uncompressed MLP weights represent a significant fraction(${\sim}25\%$) of the total size. We hypothesize that at lower bit rates, compression of MLPs can achieve further reduction in model size. Our work focuses only on the compression of feature grids and can potentially benefit from the complimentary works on MLP compression.

We additionally outperform JPEG for the full range of BPP with a much larger gap at lower bitrates. Notice the blocking artifacts in \cref{fig:kodak_qualitative} for JPEG while we achieve consistent reconstructions. However, the classical methods of JPEG2000, BPG and the autoencoder based work of~\cite{balle2018variational} continue to outperform our method, especially in the higher BPP regime. Nevertheless, we reduce the gap between INRs and autoencoder in the low-dimensional image space.

In order to understand the scalability of various image compression methods with image resolution, we compress images in the UVG-F dataset ($1920\times 1080)$ and 4 images with increasing resolution. Results are summarized in Table~\ref{tab:gigapix_baselines}. We outperform~\cite{strumpler2022implicit} by a large margin on the UVG-F dataset with an improvement of over 4.5dB while also requiring $2\times$ fewer bits. This is also observed in~\cref{fig:kodak_qualitative} (right column) where we capture much finer high frequency details compared to the positional encoding approach of \cite{strumpler2022implicit}. We marginally outperform JPEG as well in the high BPP regime which again exhibits minor blocking artifacts. Perhaps, the most surprising result is the performance of the Rate-Distortion AutoEncoder (RDAE) based approach~\cite{balle2018variational} which does not scale very well to large dimensions. We obtain a 3.5dB improvement in PSNR while maintaining a similar BPP albeit with slightly lower SSIM score.

For higher image resolutions, we continue to observe negligible drops in performance compared to INGP~\cite{muller2022instant} while achieving $4{-}9\times$ smaller bitrates. We qualitatively visualize results on the Cosmic-cliffs image (with a resolution of $8441\times14575$) in~\cref{fig:cosmic}. We achieve a similar PSNR as Instant-NGP (38.78 dB) with ${\sim}6{\times}$ reduction in bitrates. MLP-based INRs such as SIRENs~\cite{sitzmann2020implicit} are unable to capture the high frequency detail resulting in blurry images and achieve low PSNR (27.32 dB). JPEGs also lead to a large drop in performance in the similar low BPP regime(${\sim}0.15$).

\begin{figure*}[t]
\centering
\vspace{-2em}
   \includegraphics[width=0.9\linewidth]{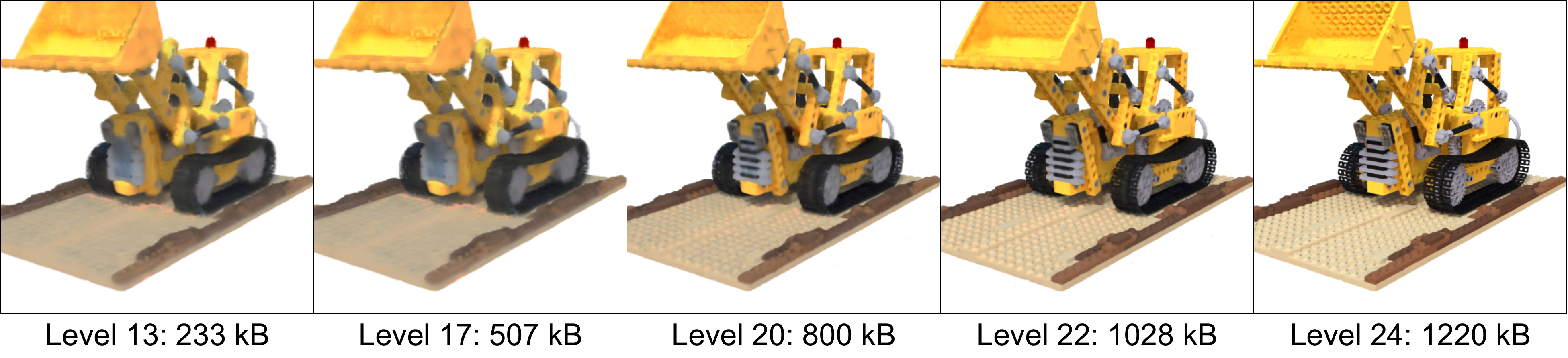}
   \caption{Our multiresolution compressed representations can be transmitted at varying LODs at inference time (without any retraining) making it suitable for applications with progressive streaming.}
\label{fig:streaming_lod}
\vspace{-1em}
\end{figure*}

\subsection{Radiance fields compression}
\label{ssec:3d_compress}
We now turn to the application of \ours to neural radiance fields or NeRFs. We compare our approach against the baseline Instant-NGP~\cite{muller2022instant}, mip-NERF~\cite{barron2021mip} and VQAD~\cite{takikawa2022variable}, a codebook based feature-grid compression approach. We evaluate on 10 brick scenes from the RTMV dataset by training each approach for 600 epochs and summarize the results in Table~\ref{tab:rtmv_baselines}. We marginally outperform the baseline INGP in terms of all the reconstruction metrics of PSNR, SSIM and LPIPS while also achieving ${\sim}48\times$ compression requiring an average of only 1MB per scene. We also outperform mip-NERF performing better on PSNR and SSIM while reducing the storage size. For a better comparison with VQAD (based off of NGLOD-NERF~\cite{takikawa2021neural}), we scale down T, the maximum number of hashes. We see that we obtain $>20\%$ lower model size at 0.43 MB compared to 0.55 MB of VQAD while slightly better in the reconstruction metrics. We also see a clear improvement over VQAD for the LEGO NeRF scene visualized in ~\Cref{fig:teaser}. VQAD has around 1.5dB PSNR drop while still being ${\sim}7{\times}$ larger in terms of storage size. Additionally, VQAD fails to scale for higher bitwidth due to memory constraints even on an NVIDIA RTX A100 GPU with 40GB GPU RAM.

To better illustrate the 3 approaches of INGP, VQAD and \ours, we train on the full resolution V8 scene ($1600~\times~1600$) for 600 epochs, visualizing the results in~\cref{fig:v8_qualitative}. We outperform VQAD (+1dB) at similar model size and obtain ${\sim}60\times$ compression compared to INGP, but with 1dB drop in PSNR. Nevertheless, we reconstruct the scene with fewer artifacts and consistency in the intricate shapes, compared to VQAD as highlighted in the patches.

\renewcommand{\arraystretch}{1.1}
\begin{table}[t]
\vspace{0.5em}
\caption{Comparison against various video INR approaches on the UVG dataset. We outperform NIRVANA with higher PSNR and lower BPP while obtaining slightly lower PSNR than NeRV at $3\times$ reduction in model size.}
\centering
\resizebox{\linewidth}{!}{
\begin{tabular}{@{}L{\dimexpr.15\linewidth}C{\dimexpr.34\linewidth}C{\dimexpr.15\linewidth}C{\dimexpr.12\linewidth}@{}}
\toprule
Method & Encoding Time $\downarrow$ &PSNR $\uparrow$  & BPP $\downarrow$ \\
\midrule
SIREN & 15 hours & 27.20  & 0.28 \\
NeRV & 3.5 hours & \textbf{35.54} &0.66\\
NIRVANA & 4 hours & 34.71 & 0.32 \\
Ours & 6.5 hours & 35.01 & \textbf{0.21} \\
\bottomrule
\end{tabular}
} %
\vspace{-1em}
\label{tab:uvg_baselines}
\end{table}

\renewcommand{\arraystretch}{1.1}
\begin{table}[t]

\caption{Comparison of various methods on scenes in the RTMV dataset. We marginally outperform the baseline Instant-NGP in terms of all metrics while also achieving a $48\times$ compression.}

\centering
\resizebox{\linewidth}{!}{
\begin{tabular}{@{}L{\dimexpr.3\linewidth}C{\dimexpr.14\linewidth}C{\dimexpr.13\linewidth}C{\dimexpr.15\linewidth}C{\dimexpr.2\linewidth}@{}}
\toprule
Method & PSNR $\uparrow$ & SSIM $\uparrow$& LPIPS $\downarrow$ & Storage $\downarrow$ \\
\midrule
NeRF&28.28&0.9398&0.0410&2.5MB\\
mip-NERF & \textbf{31.61} & \textbf{0.9582} & \textbf{0.0214} & \textbf{1.2MB} \\
\midrule
NGLOD-NERF & \textbf{32.72} & \textbf{0.9700} & \textbf{0.0379} & $\approx$20MB\\
VQAD & 31.45 & 0.9638 & 0.0468 & 0.55MB\\
Ours & 31.46 & 0.9657 & 0.0428 & \textbf{0.43MB}\\
\midrule
Instant-NGP&31.88&0.9690&0.0381&48.9MB\\
Ours & \textbf{32.14} & \textbf{0.9704} & \textbf{0.0348} & \textbf{1.03MB}\\
\bottomrule
\end{tabular}
} %
\vspace{-1em}
\label{tab:rtmv_baselines}
\end{table}

\subsection{Video compression}
\label{ssec:video_compress}
Next, we apply our approach to video compression as well. As a baseline, we compare against SIREN a coordinate-based INR. We also compare against NeRV~\cite{chen2021nerv}, a video INR based approach which takes a positional encoding as input and predicts frame-wise outputs for a  video. We also compare against another video INR, NIRVANA~\cite{maiya2022nirvana}, which is an autoregressive patch-wise prediction framework for compressing videos. We run the baselines on the UVG dataset, with the results shown in Table~\ref{tab:uvg_baselines}. 

We outperform SIREN by a significant margin with an almost $+7dB$ gain in PSNR and $25\%$ lesser BPP and shorter encoding times. This is to be expected as SIREN fails to scale to higher dimensional signals such as videos usually with more than $10^9$ pixels. We also outperform NIRVANA achieving higher PSNR and lower BPP albeit at longer encoding times. Compared to NeRV, we obtain a 0.5dB PSNR drop but achieve $3{\times}$ compression in model size. We would like to add that our current implementation utilizes PyTorch and the encoding time can be reduced significantly using fully fused CUDA implementation (\cite{muller2022instant} demonstrated that efficient lookup of the hashtables can be an order of magnitude faster than a vanilla Pytorch version). Additionally, our approach is orthogonal to both baselines and provides room for potential improvements. For instance, compressed multi-resolution feature grids can be used to replace positional embedding for NeRV as well as coordinate-based SIREN, and provide faster convergence and better reconstruction for high dimensional signals.

\subsection{Streaming LOD}
\label{ssec:streaming}
\cite{takikawa2022variable} show the advantage of feature-grid based INRs for progressive streaming at inference time due to their multi-resolution representation capabilities. Since our compression framework consists of latents at different LODs as well, they can be progressively compressed with varying LOD yielding reconstructions with different resolution. Formally, for $\Qb~{=}~\{\Qb_1,\Qb_2,...\Qb_L\}$ we can reconstruct the signal at LOD $l$ by only passing the first $l$ latents while masking out the finer resolutions. This can be applied directly during inference without any additional retraining. We visualize the effect of such progressive streaming in~\cref{fig:streaming_lod}. We obtain better reconstruction with increasing bitrates or the latent size. This is especially beneficial for streaming applications as well as for easily scaling the model size based on the storage or bandwidth constraints.

\begin{figure}[t]
   \centering
   \includegraphics[width=0.95\linewidth]{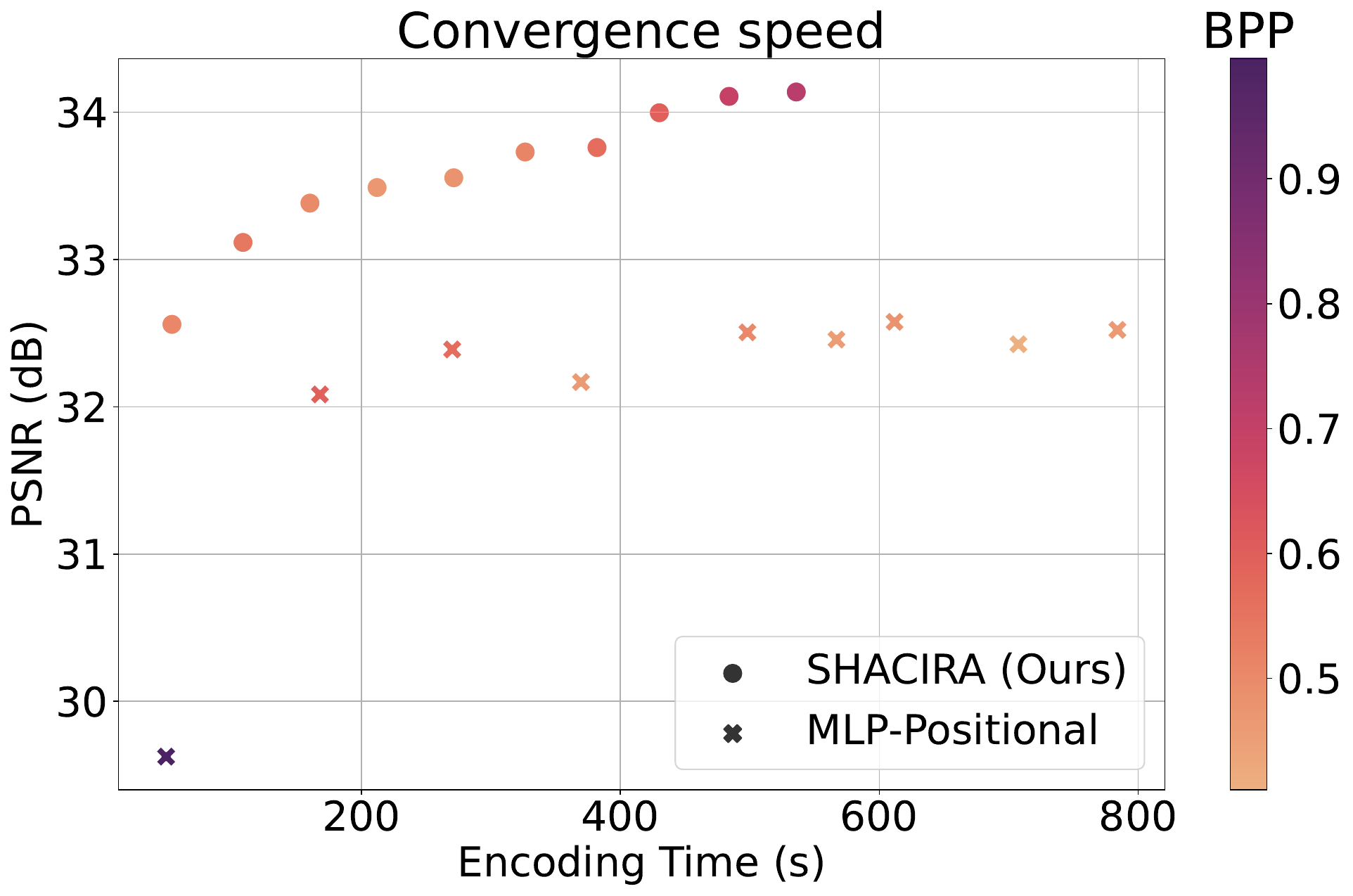}
   \vspace{-1em}
   \caption{Comparison of convergence speeds of our feature-grid approach with~\cite{strumpler2022implicit} utilizing MLP-based INRs. We obtain higher PSNR at a much faster rate for a similar BPP range.}
   \vspace*{-1em}
\label{fig:convergence}
\end{figure}

\subsection{Convergence speeds}
\label{ssec:convergence}
We now compare the convergence speeds of our feature-grid based approach with that of~\cite{strumpler2022implicit} which is an MLP-based INR with positional encoding. We summarize the results for an image in the Kodak dataset in~\Fig{\ref{fig:convergence}}. We measure encoding times on an NVIDIA RTX A6000 GPU for the full length of the training for both approaches. We obtain higher PSNR at a much faster rate than~\cite{strumpler2022implicit} at a similar BPP range (hue value in color bar). While~\cite{strumpler2022implicit} reduces BPP (0.84 at 600s) with higher encoding times, their PSNR remains stagnant at 32.5dB. In contrast, our approach achieves this PSNR and BPP (0.85) within just 53s achieving more than a $10\times$ speedup in convergence. Additionally, we achieve higher PSNR with longer encoding times reaching 34dB in 430s while maintaining a similar BPP (0.87).

\subsection{Effect of entropy regularization and annealing}
\label{ssec:ent_anneal}
In this section, we analyze the effect of entropy regularization and annealing. We pick the Jockey image from UVG-F ($1080\times~1920$) for our analysis. We set the default values of latent and feature dimensions to $1$. For the analysis, we compare trade-off curves by increasing the number of entries from $2^{13}$ to $2^{17}$ in multiples of 2 which naturally increases the number of parameters and also the PSNR and BPP. Note that better trade-off curves indicate shifting upwards (higher PSNR) and to the left (lower BPP).

\begin{figure}[t]
   \centering
   \includegraphics[width=0.93\linewidth]{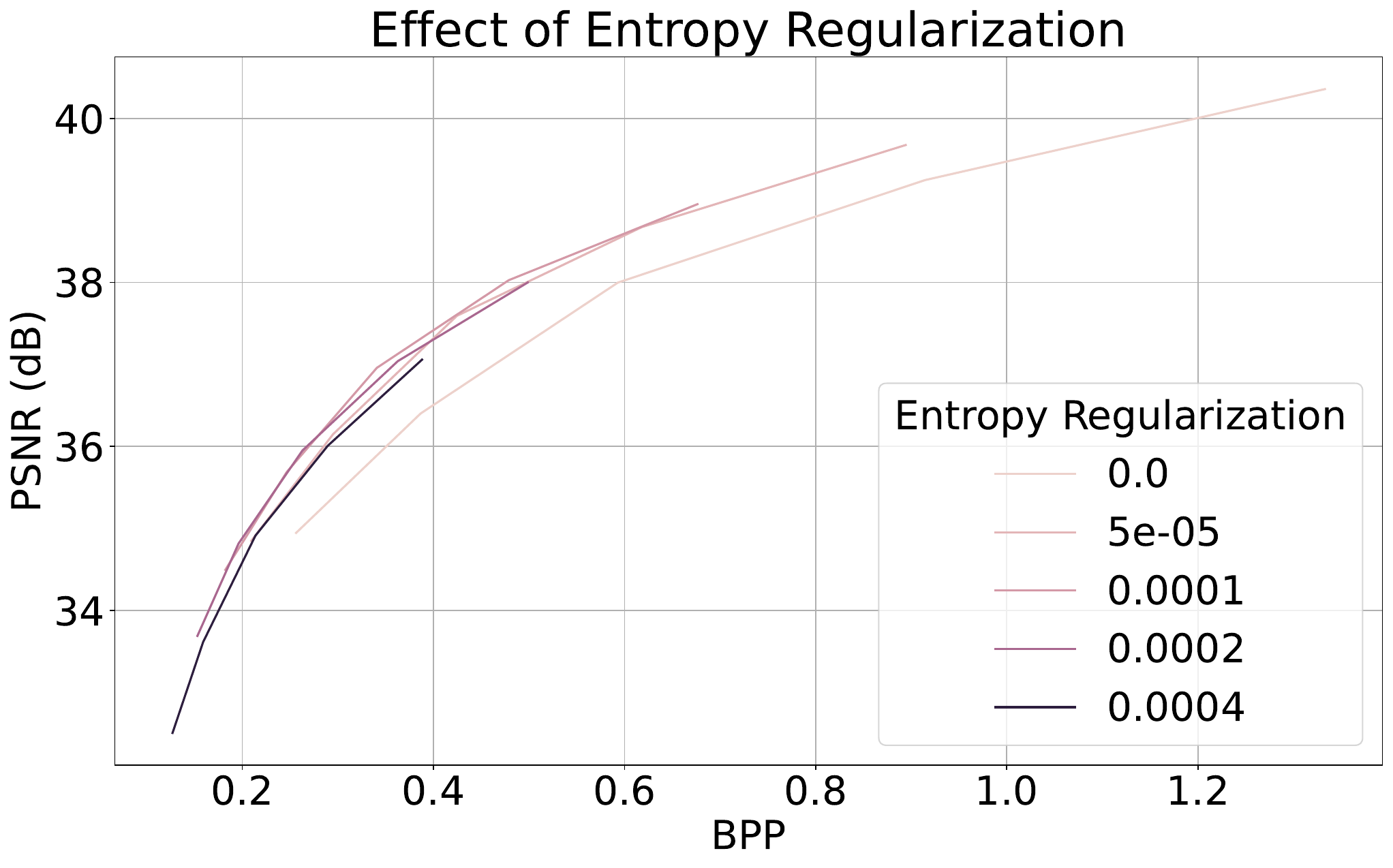}
   \vspace{-1em}
   \caption{Effect of entropy regularization. In absence of entropy regularization, corresponding to the value $0.0$, there is a drop in the PSNR-BPP trade-off curve.}
\label{fig:abl_ent}
\end{figure}

~\Cref{fig:abl_ent} shows the effect of entropy regularization. The absence of it, corresponding to a value of $0.0$ shows a drop in performance compared to values of $1.0e^{-4}$ and higher. We see that the network performance is fairly stable in this range with much higher values of $4.0e^{-4}$ showing small drops. This shows that entropy regularization using the defined probability models helps in improving the PSNR-BPP trade-off curve by decreasing entropy (or model size/BPP) with no drop in network performance (PSNR).

\begin{figure}[t]
   \centering
   \includegraphics[width=0.93\linewidth]{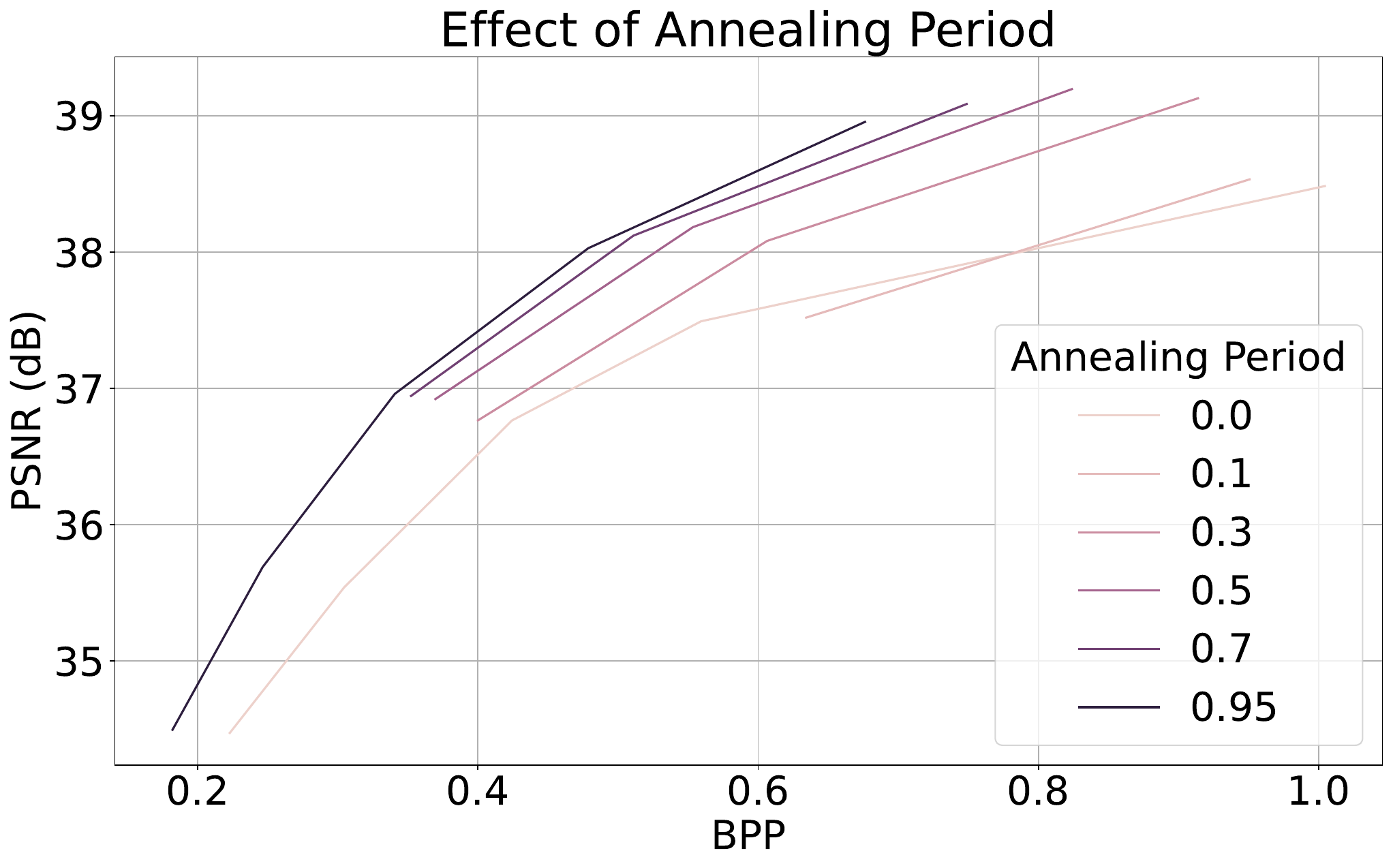}
   \vspace{-1em}
   \caption{Effect of annealing. Increasing the fractional period of annealing improves the trade-off curve highlighting its importance for quantization approximation compared to stand-alone STE (Straight Through Estimator) with an annealing period of 0.0.}
\label{fig:abl_annealing}
   \vspace{-0.5em}
\end{figure}

~\Cref{fig:abl_annealing} shows the effect of annealing. We vary the duration of annealing as a fraction of the total training duration. No annealing corresponds to the value $0.0$ and has a large drop in the trade-off curve compared to higher values. This shows that annealing performs better than standard STE (Straight Through Estimator) alone and is an important component of our approach for quantizing the feature grid parameters. Increasing the period of annealing shows consistent improvements in the trade-off curve.

\section{Conclusion}
\label{sec:conclusion}

We proposed \ours, a general purpose, end-to-end trainable framework for learning compressed INRs or neural fields for images, videos, and NeRFs. Our method adds a minimal overhead over feature-grid based methods of training INRs. We make this possible by parameterizing the feature grid of INRs with discrete latents and a decoder. Applying entropy regularization on discrete latents ensures that we can learn and maintain compressed quantized weights. A tiny decoder allows us to still operate in the continuous space for training and sampling with high fidelity. Our method requires no posthoc training or quantization and can be easily added to existing INR pipelines. We conduct an extensive evaluation on images, videos, and 3D datasets to show the efficacy of our approach in these domains. Compared to MLP-based INRs, we scale well for high resolution signals capturing high frequency details and converge faster. An additional benefit of our approach is that it allows reconstruction of the signal at different levels-of-details without retraining, which is especially beneficial for streaming applications. While the compressed network is memory efficient, it does not offer any inference speedups compared to the uncompressed network and is an interesting avenue for further research. Meta learning the feature grid for even faster convergence is another direction for future work.

\clearpage
{\small
\bibliographystyle{ieee_fullname}
\bibliography{egbib}
}
\clearpage

\appendix
\appendixpage
\ificcvfinal\thispagestyle{empty}\fi
\section{Probability models}
We define the probability models similar to~\cite{balle2018variational}. The underlying probability density of the latents $\Qhb$ is defined by its Cumulative Density Function (CDF) $c: \mathbb{R}\rightarrow[0,1]$, with the constraints:
\begin{equation}
    c(-\infty)=0, \quad c(\infty)=1, \quad \frac{\partial c(x)}{\partial x} \geq 0
\end{equation}
This is represented using MLPs which take in a real valued scalar and output a CDF value between 0 and 1. Each dimension in $\Qhb$ is represented by a separate model. To satisfy the monotonicity constraint,~\cite{balle2018variational} use a combination of tanh and softplus activations for each layer of the MLP. A sigmoid activation is used at the final layer to constrain the CDF between 0 and 1. To model the true distribution, the standard uniform distribution $n\sim\mathcal{U}[-0.5,0.5]$ is convolved with the density model to derive the Probability Mass Function (PMF) of the latents as 
\begin{equation}
    P_d(x) = c\left(\Qhb_d+\frac{1}{2}\right)-c\left(\Qhb_d-\frac{1}{2}\right)
\end{equation}
The entropy regularization loss is then the self information loss given by 

\begin{equation}
\mathcal{L}_I(\Qhb) = -\frac{1}{T}\sum_{d=1}^{D}\sum_{i=1}^{T}\textnormal{log}_2\left(P_d(x)\right)
\end{equation}

\section{Experimental settings}
The latent dimension is set to 1 for images and video and 2 for 3D experiments. The feature dimension obtained after decoding the latents is set to 1 for Kodak and UVG-F, 2 for high resolution (giga-pixel) images and UVG for videos, and 4 for 3D experiments.
Since we do not compress MLP weights but include their floating-point size in our PSNR-BPP tradeoff calculation, we vary the hidden dimension according to the signal resolution. For images, we set the dimension to be 16 for Kodak, 48 for UVG-F and SMACS, and 96 for the remaining high-resolution images. For videos and 3D, we set the layer size to 128. The number of layers is fixed to 2 for all cases. Note that higher layer size generally leads to better PSNR at the cost of a proportionally higher BPP but further gains can be obtained by compressing these weights as well. This is orthogonal to our direction of compression of the feature grid itself.

For all our experiments, we initialize the decoder parameters with a normal distribution $\mathcal{N}(0,0.1)$, latents with $\mathcal{U}(-0.01,0.01)$, MLP weights with the Xavier initialization~\cite{glorot2010understanding}, and probability model parameters as in~\cite{balle2018variational}. 

We use the Adam optimizer for jointly optimizing all network parameters. We set the learning rate of MLP parameters to be $1.0e^{-3}$ for Kodak and $5.0e^{-4}$ for all other experiments. The learning rate for the probability models is fixed at $1.0e^{-4}$ following~\cite{Girish2022LilNetXLN}. The decoder learning rate is set to $0.01$ for images, video and $0.1$ for 3D experiments. We set the learning rate of the latents to be $0.01$ for Kodak, UVG-F, and UVG, $0.05$ for 3D, and $0.1$ for the other higher-resolution images.
We observe that the training is not very susceptible to variations in the initialization or learning rates and the values are obtained from a coarse search for each signal domain.

\section{Sensitivity Analysis}
In this section, we analyze the effect of various components of our pipeline. We pick the Jockey image ($1080\times1920$) from UVG-F as the benchmark for our analysis. We set the default values of entropy regularization to $1.0e^{-4}$, latent and feature dimensions to $1$ each, MLP width to $48$, MLP depth to $2$, and annealing period to $0.95$. Each subsection below analyzes varying a single parameter while keeping others fixed to their default values. For the analysis, we compare tradeoff curves by increasing the number of entries from $2^{13}$ to $2^{17}$ in multiples of 2 which provides a natural way of increasing the number of parameters and subsequently the PSNR and BPP. Note that better tradeoff curves indicate shifting upwards (higher PSNR) and to the left (lower BPP).

\begin{figure*}[t]
\centering
\setlength{\tabcolsep}{10pt}
\begin{tabular}{ccc}
   \includegraphics[width=0.3\linewidth]{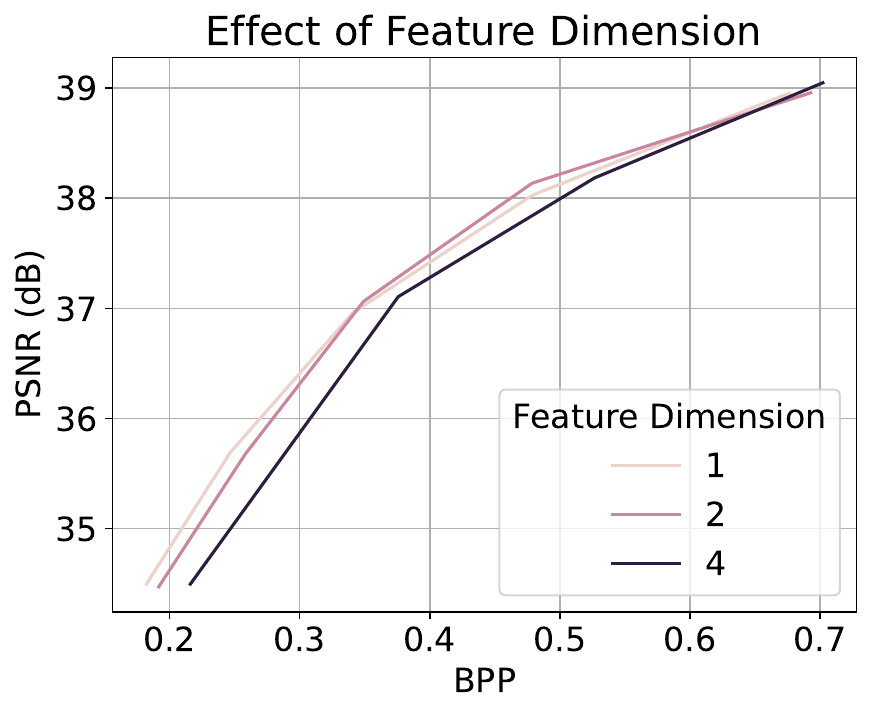}&
   \includegraphics[width=0.3\linewidth]{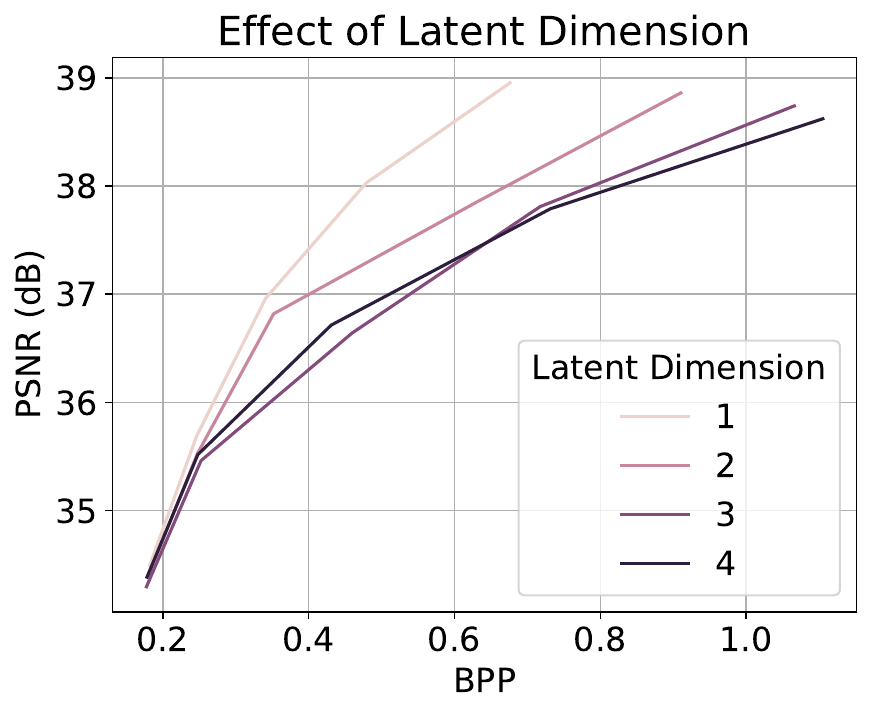}&
   \includegraphics[width=0.3\linewidth]{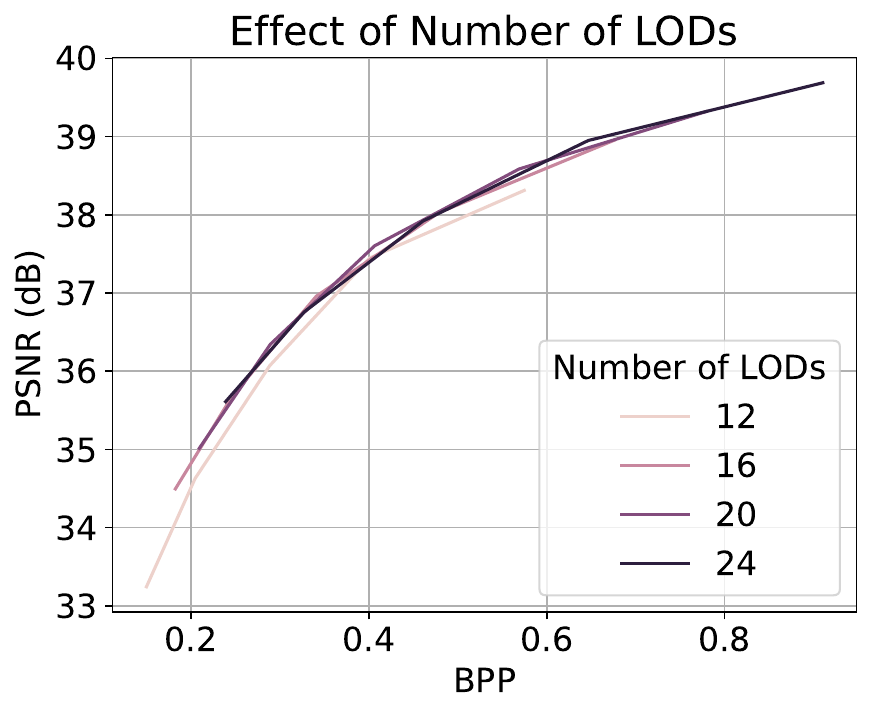}\\
   \quad\quad (a)&\quad\quad(b)&\quad\quad(c)
\end{tabular}
   \caption{Effect of latent and feature dimension and number of LODs. Increasing the latent dimension leads to a much higher size at smaller increases in PSNR. Optimal value of feature dimension (or decoded latents) is 1 or 2. Increasing the number of LODs shifts the curve upwards and to the right yielding no difference in tradeoff curves.}
\label{fig_supp:abl_FTL}
\end{figure*}

\begin{figure}[t]
\centering
\setlength{\tabcolsep}{0pt}
\begin{tabular}{cc}
   \includegraphics[width=0.5\linewidth]{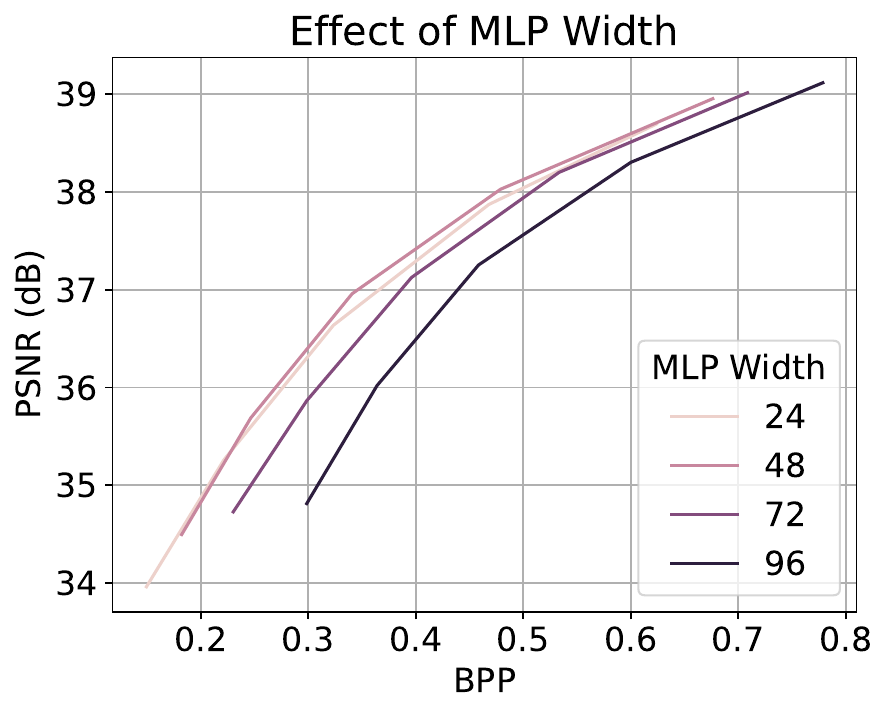}&
   \includegraphics[width=0.5\linewidth]{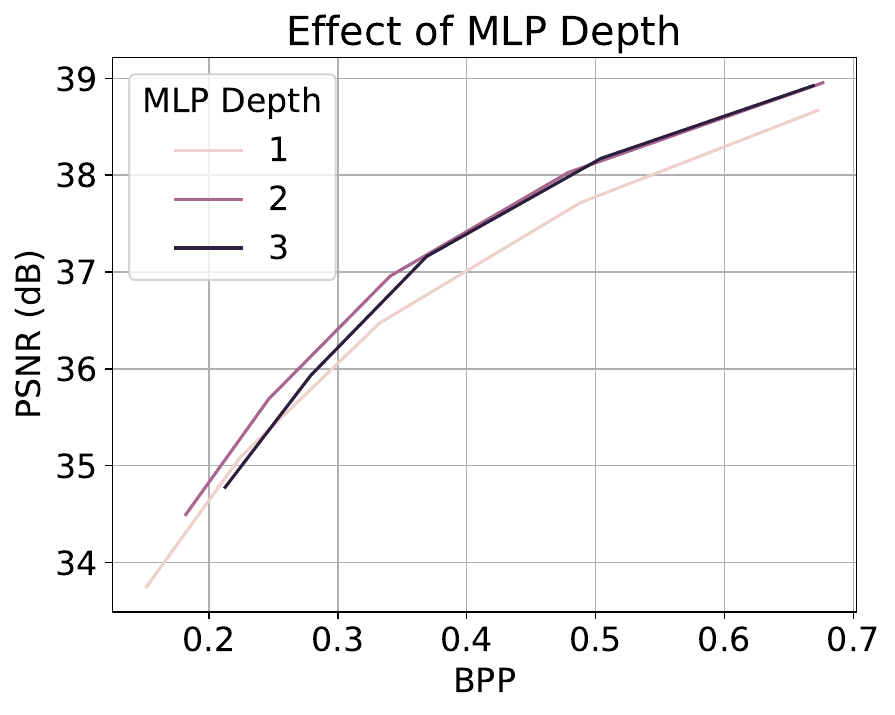}\\
   \quad\quad (a)&\quad\quad(b)
\end{tabular}
   \caption{Effect of MLP Width and Depth. Increasing MLP hidden layer size (width) is detrimental due to a large increase in uncompressed parameters (and BPP) with a small increase in PSNR. Increasing the number of hidden layers beyond 2 has little effect on the tradeoff curve.}
\label{fig_supp:abl_mlp}
\end{figure}

\subsection{Effect of Latent and Feature Dimension}
We ablate the number of LODs, the number of latent dimensions, and feature dimensions after decoding the latents. Results are shown in~\Fig{\ref{fig_supp:abl_FTL}}. We see that increasing the feature dimension from 1 to 2 does not significantly alter the tradeoff curve while increasing it to 4 leads to a small drop in performance. On the other hand, increasing the latent dimension has a strong impact on the tradeoff curve as it directly impacts the number of latents entropy encoded. BPP visibly increases with higher latent dimension but leads to only modest gains in PSNR. Increasing the number of LODs shifts the curve upwards and to the right but has no overall impact in terms of improving PSNR for a fixed BPP.

\subsection{Effect of MLP Width and Depth}
Finally, we analyze the effect of the hidden dimension (width) of the MLP as well as the number of layers (depth) in~\Fig{\ref{fig_supp:abl_mlp}}. Increasing the depth from 1 to 2 shows a marginal improvement while further increases do not have a large effect. This shows that the MLP's representation capability caps at a certain value as the majority of parameters are present in the feature grid. Increasing MLP width on the hand leads to a clear drop in performance as the number of uncompressed parameters in the MLP increases approximately quadratically leading to a larger BPP (shifting to the right) but with only small gains in PSNR (small shift up).

\begin{figure}[h]
   \includegraphics[width=\linewidth]{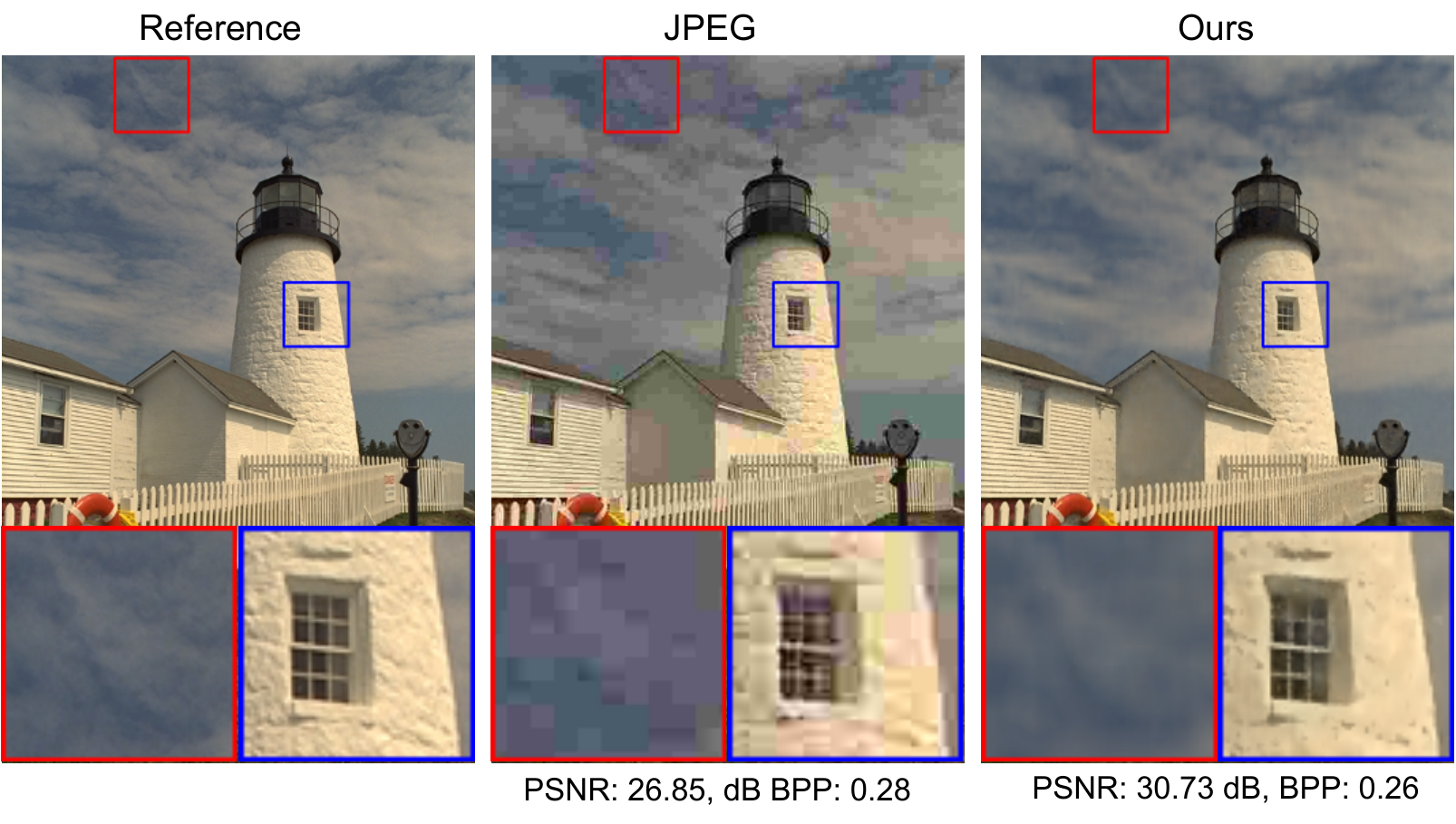}
   \vspace{-2em}
   \caption{Qualitative results on Kodak. We significantly outperform JPEG in the low BPP regime (${+}4dB$) at similar BPP~(${\sim}0.26$). JPEG fails to capture the intricate details in the image such as the window of the building or the clouds in the sky, exhibiting severe artifacts.}
\label{fig_supp:kodak_lighthouse}
\end{figure}

\section{Feature grid visualization}

\begin{figure*}
   \includegraphics[width=\linewidth]{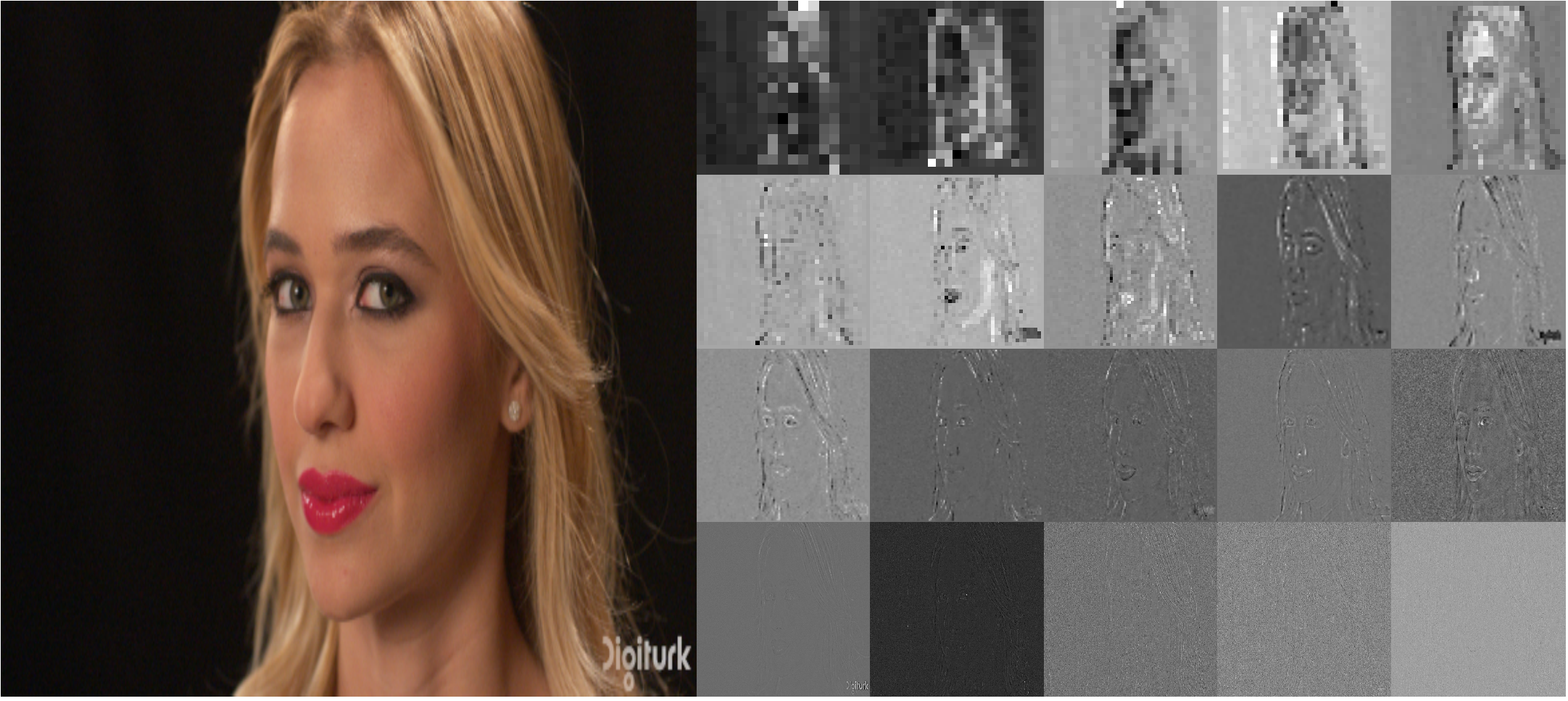}
   \vspace{-2em}
   \caption{Visualization of learned features (right) for Beauty image (left) from UVG-F. Feature maps at increasing LODs (left-to-right then top-to-bottom) capture finer details in the image highlighting the hierarchical features learned for each signal.}
\label{fig_supp:feat_viz}
\end{figure*}

\begin{figure*}[t]
   \includegraphics[width=\linewidth]{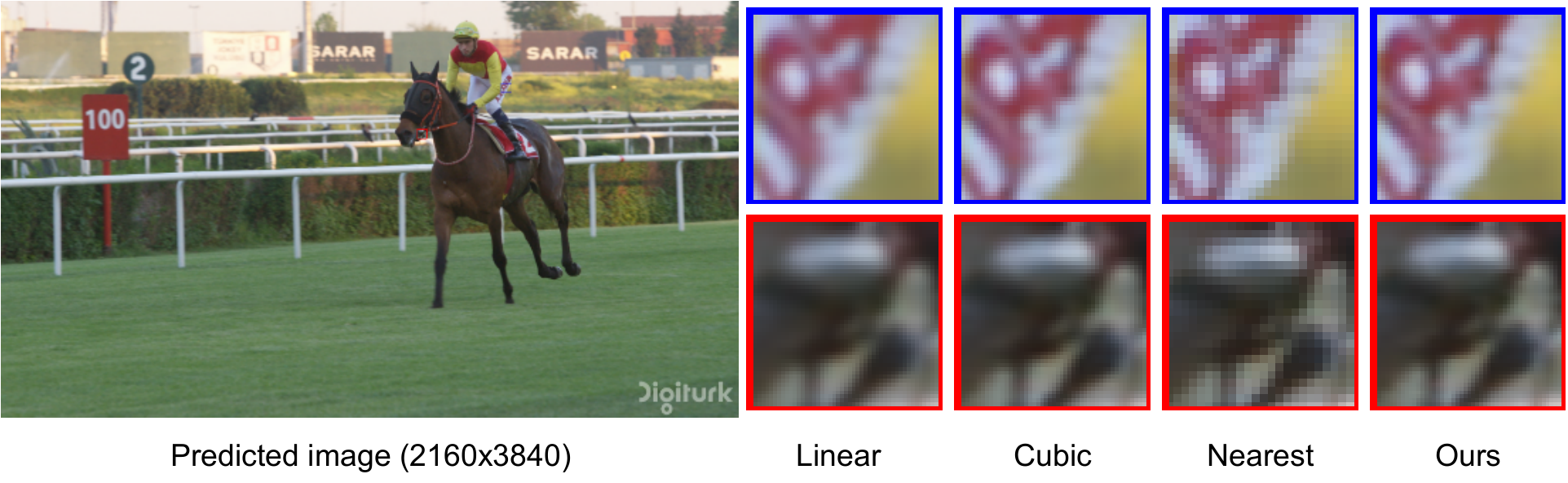}
   \vspace{-2em}
   \caption{Application of our approach for super-resolution of the Jockey image in the UVG-F dataset. We obtain sharper images compared with standard upsampling methods at 2x upsampling factor ($2160\times3840$).}
\label{fig_supp:superres}
\end{figure*}

We visualize the learned latents after training on the Beauty image from UVG-F in~\Fig{\ref{fig_supp:feat_viz}. With increasing LOD or feature resolution (from left-to-right and top-to-bottom), we see that finer details of the image are captured. Thus, the latents represent the image features at different scales/levels. This can be particularly useful for downstream tasks which may require features at different scales. Additionally, such a hierarchy enables the application of our method in streaming scenarios with higher bitrates leading to higher PSNR (as also discussed in Sec.4.5). Beyond a certain level, we see that the features become less informative globally. This is due to the fact that the grid resolution at finer levels is larger than the number of entries in the latents which is fixed. Multiple locations in the grid map to the same entry in the latent space.

\section{Image superresolution}

We show the capability of our approach to perform super-resolution of images by providing the input coordinate grids at the desired resolution. \Fig{\ref{fig_supp:superres}} shows results on the super-resolution of the Jockey image from UVG-F ($1080\times1920$ resolution) by a factor of 2. We obtain slightly sharper images as compared to standard upsampling methods such as linear, cubic or nearest neighbor interpolation.

\section{Additional visualizations}

We show additional visualizations of the Pearl ($23466\times20000$) and SMACS images ($4630\times4537$) in Fig.\ref{fig_supp:pearl} (top row and bottom row respectively). We see that we continue to obtain high quality reconstructions and achieve similar PSNR compared to Instant-NGP while being much smaller in storage size (in terms of BPP). Significant artifacts and discoloration are also visible for the traditional JPEG while still requiring more memory than our approach.
\begin{figure*}[t]
   \includegraphics[width=\linewidth]{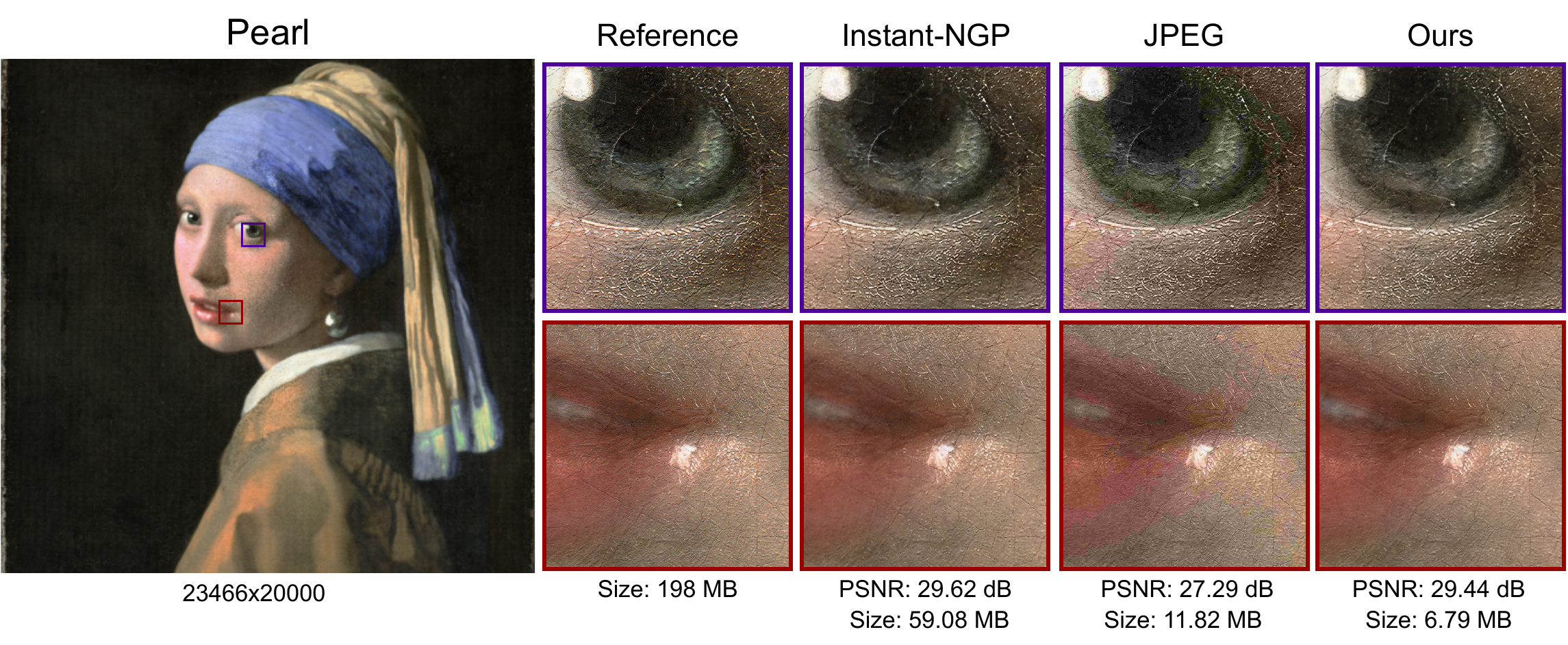}
   \includegraphics[width=\linewidth]{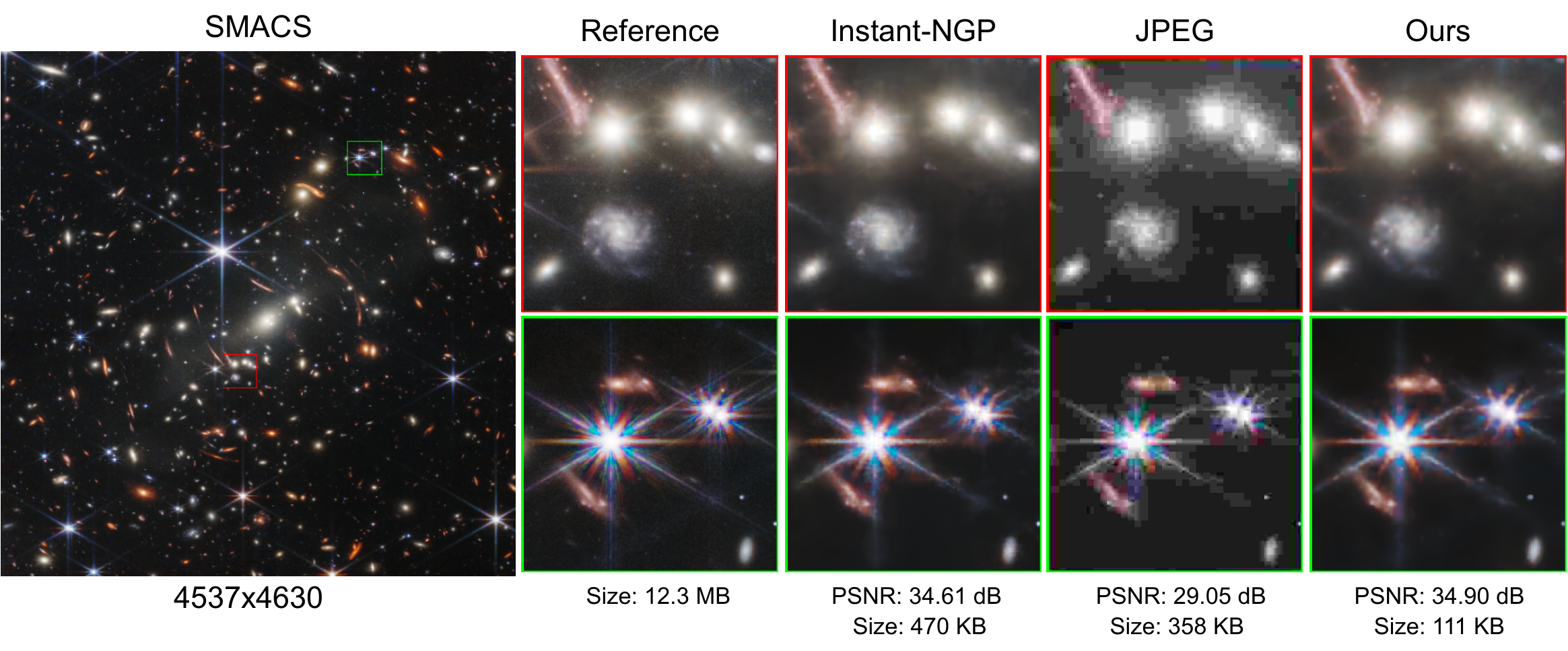}
   \vspace{-2em}
   \caption{Visualization of Pearl and SMACS images. For Pearl, we maintain a similar PSNR as Instant-NGP~\cite{muller2022instant} while being ${\sim}9{\times}$ smaller. For SMACS, we marginally outperform them with more than a $4{\times}$ compression factor. JPEG exhibits visual discoloration artifacts leading to a much lower PSNR even with lower compression factors.}
\label{fig_supp:pearl}
\end{figure*}

\end{document}